\documentclass{article}

\newif\ifanonymous
\newif\ifshort
\shortfalse

\usepackage{microtype}
\usepackage{graphicx}
\usepackage{subfigure}
\usepackage{booktabs} 
\usepackage{xcolor} 
\usepackage{svg}
\usepackage{adjustbox}
\usepackage{multirow}
\usepackage{dejavu}
\usepackage{placeins}

\usepackage{hyperref}
\usepackage{siunitx}

\usepackage[accepted]{icml2025}

\usepackage{amsmath}
\usepackage{amssymb}
\usepackage{mathtools}
\usepackage{amsthm}
\usepackage{bm}
\usepackage{nicefrac}
\usepackage{listings}
\usepackage{enumitem}
\setlist[enumerate]{itemsep=-2pt,topsep=-2pt}

\usepackage[capitalize]{cleveref}
\Crefname{equation}{Eq.}{Eqs.}
\Crefname{figure}{Fig.}{Figs.}
\Crefname{tabular}{Tab.}{Tabs.}
\Crefname{appendix}{App.}{Apps.}
\Crefname{section}{Sec.}{Secs.}

\theoremstyle{plain}

\theoremstyle{definition}

\theoremstyle{remark}

\usepackage[textsize=tiny]{todonotes}

\definecolor{OurRed}{RGB}{240, 60, 30}
\definecolor{NavyBlue}{RGB}{0, 110, 184}
\definecolor{VioletBlue}{RGB}{120, 85, 170}
\definecolor{OurGreen}{RGB}{112, 185, 111}
\definecolor{MainFigYellow}{RGB}{227, 116, 0}
\definecolor{MainFigGreen}{RGB}{52, 168, 83}

\newcommand{\stress}[1]{{\fontfamily{LinuxLibertineT-OsF}\fontseries{sb}\selectfont #1}}

\newcommand{\gemma}{\textsc{Gemma2}}

\newcommand{\gist}{\textsc{Gist}}
\newcommand{\offsetgist}{\textsc{OffsetGist}}
\newcommand{\sepoffsetgist}{\textsc{SepOffsetGist}}
\newcommand{\newgist}{\textsc{GistPool}}
\newcommand{\avgpool}{\textsc{AvgPool}}
\newcommand{\geminijudge}{\textsc{Gemini Judge}}
\newcommand{\geminicompress}{\textsc{Gemini Compress}}

\newcommand{\alpaca}{\textsc{Alpaca+}}
\newcommand{\squad}{\textsc{SQuAD}}
\newcommand{\drop}{\textsc{DROP}}
\newcommand{\race}{\textsc{RACE}}
\newcommand{\fairytale}{\textsc{FairytaleQA}}
\newcommand{\narrative}{\textsc{NarrativeQA}}

\newcommand{\fullbaseline}{\stress{Full context baseline}}
\newcommand{\nobaseline}{\stress{No context baseline}}

\newcommand{\losslessT}{\hyperlink{tar_lossless_t}{\stress{lossless transition}}}
\newcommand{\scalability}{\hyperlink{tar_scalability}{\stress{scalability}}}

\newcommand{\archiconst}{\hyperlink{tar_archi_const}{\stress{architectural consistency}}}

\newcommand{\sphereemb}{\stress{hypersphere embeddings}}
\newcommand{\gemmaemb}{\stress{Gemma embeddings}}
\newcommand{\fixedcontext}{\stress{fixed context size}}
\newcommand{\varcontext}{\stress{variable context size}}
\newcommand{\Standardmask}{\stress{Standard mask}}
\newcommand{\standardmask}{\stress{standard mask}}
\newcommand{\Poolmask}{\stress{Pool mask}}
\newcommand{\poolmask}{\stress{pool mask}}

\newcommand{\Xc}{C}

\newcommand{\Xg}{G}

\newcommand{\gdm}{Google DeepMind}

\icmltitlerunning{Long Context In-Context Compression by Getting to the Gist of Gisting}

\begin{document}

\twocolumn[
\icmltitle{
Long Context In-Context Compression by Getting to the Gist of Gisting
}

\icmlsetsymbol{equal}{*}

\begin{icmlauthorlist}
\icmlauthor{Aleksandar Petrov\textsuperscript{\textdagger}}{gdm,ox,equal}
\icmlauthor{Mark Sandler}{gdm}
\icmlauthor{Andrey Zhmoginov}{gdm}
\icmlauthor{Nolan Miller}{gdm}
\icmlauthor{Max Vladymyrov}{gdm,equal}
\end{icmlauthorlist}

\icmlaffiliation{gdm}{\gdm}
\icmlaffiliation{ox}{University of Oxford}

\icmlcorrespondingauthor{Max Vladymyrov}{mxv@google.com}

\icmlkeywords{Machine Learning, ICML}

\vskip 0.3in
]

\printAffiliationsAndNotice{\icmlEqualContribution} %

\begin{abstract}
Long context processing is critical for the adoption of LLMs, but existing methods often introduce architectural complexity that hinders their practical adoption.
Gisting, an in-context compression method with no architectural modification to the decoder transformer, is a promising approach due to its simplicity and compatibility with existing frameworks.
While effective for short instructions, we demonstrate that gisting struggles with longer contexts, with significant performance drops even at minimal compression rates.
Surprisingly, a simple average pooling baseline consistently outperforms gisting.
We analyze the limitations of gisting, including information flow interruptions, capacity limitations and the inability to restrict its attention to subsets of the context.
Motivated by theoretical insights into the performance gap between gisting and average pooling, and supported by extensive experimentation, we propose GistPool, a new in-context compression method.
GistPool preserves the simplicity of gisting, while significantly boosting its performance on long context compression tasks.
\end{abstract}

\section{Introduction}
\label{introduction}

The rapid adoption of Large Language Models (LLMs) necessitates ever-longer context windows to process increasing information volumes.
Current state-of-the-art models, while supporting substantial context lengths, are insufficient for emerging use-cases like web browsing agents (analyzing full HTML pages and histories), personalized assistants (requiring comprehensive user interaction records), and coding assistants (accessing extensive codebases).
Furthermore, reasoning agents and test-time inference may require hundreds of thousands, even millions, of tokens for complex reasoning.
Longer context improves reasoning and in-context learning, with smaller models being comparable  to larger models with shorter contexts~\citep{chung2024scaling}.

Multimodal models further exacerbate this challenge due to lengthy video and audio inputs.
The high cost of processing extended contexts, compounded by declining model effectiveness with long sequences \citep{liu2024lost,barbero2024transformers}, highlights the importance of efficient long-context processing for scalable LLM deployment, particularly for personalized models serving billions of users.

We focus on the decoder-only transformer architecture, given its role in most leading models.
Broadly, there are two main strategies for efficient long-context processing: introducing sparsity in the attention mechanism, or compressing the processed KV-cache.
While a comprehensive solution would ultimately integrate both strategies, this paper concentrates on the latter aspect: context compression.

For a solution to be practical at scale, simplicity and practicality are key, meaning it should have:
\begin{enumerate}
    \item \hypertarget{tar_archi_const}{\textbf{Architectural consistency:}} it should require minimal, if any, architectural modifications, ensuring seamless compatibility with existing frameworks, libraries, hardware infrastructure, and established system designs;
    \item \hypertarget{tar_scalability}{\textbf{Scalability:}} it should have similar performance across different sequence lengths;
    \item \hypertarget{tar_lossless_t}{\textbf{Lossless transition:}} its performance should gradually increase as the compression rate decreases, with the $1\times$ case recovering the base model performance. 
\end{enumerate}

Driven by this aspiration for simplicity, the concept of ``gist tokens''~\citep{mu2024learning} emerges as a simple, compelling approach to context compression.
By introducing a representational bottleneck in the attention mask, gisting forces the model to condense input information into dedicated token activations.
This generates a context-specific ``tuning prefix''~\citep{li2021prefix} in a single forward pass.
Implementation is straightforward, requiring only new vocabulary token embeddings and an attention mask modification.
Gist tokens require no model surgery, work across modalities, leverage core transformer properties, and compress in a single pass, thus possessing \archiconst{}.

Unfortunately, the original gisting approach does not meet the \scalability{} requirement in its proposed form.
We find that gisting does not effectively scale even to contexts of 100s of tokens which is relatively short for contemporary standards.
Worse, on certain datasets, performance nears random guessing even at a $1\times$ compression rate, a situation that should ideally be near lossless.
Consequently, gisting also violates the \losslessT{} property.

Although gisting underperforms, its simplicity and elegance remain appealing.
We investigate why it struggles with medium to long contexts, identify several performance issues and propose solutions.
As a result, we introduce \mbox{\newgist{}} which retains the \archiconst{} of gisting while significantly improving its \scalability{} and \losslessT{}.
Concretely, our contributions are:
\begin{enumerate}
\item We show that the original \gist{} method is effective only in scenarios with short context.
\item We demonstrate that, surprisingly, a simpler average pooling approach is more effective for longer contexts.
\item Inspired by average pooling's performance, we propose \newgist{}, a novel method improving \gist{} by: (1)~uniformly distributing tokens across the context, (2)~separately fine-tuning for gist tokens, and (3)~shifting activations down one level during the prediction phase.
\item We offer comprehensive theoretical justifications and empirical evidence that these modifications are necessary for scalable long-context in-context compression.
\end{enumerate}

\begin{figure}
    \centering
    \includegraphics[width=\columnwidth]{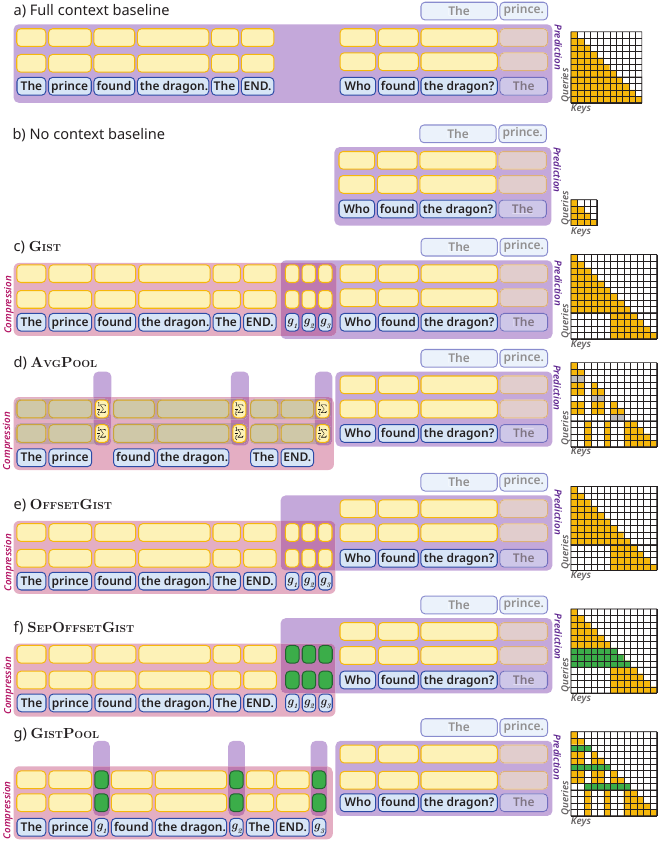}
    \vspace{-1em}
    \caption{\textbf{In-context compression methods.} We illustrate compressing a story with $\xi{=}2$ compression rate, using a 2-layer transformer. At inference time, the model first compresses the context and then autoregressively samples an answer based on the compressed context and the query. 
    Both parts are used for training.
    The color of the activations and the mask rows (\textcolor{MainFigYellow}{yellow}/\textcolor{MainFigGreen}{green}) correspond to different sets of model parameters.
    All tokens attend also to the BOS token which is not shown.
    \textbf{a)} The \fullbaseline{}, i.e., finetuning the base model.
    \textbf{b)} The \nobaseline{}, simulating the worst-case performance when the context is fully destroyed by the compression.
    \textbf{c)} The original \gist{} setup where the added gist tokens can attend to the context but the query and answer tokens can only attend to the gist tokens.
    \textbf{d)} \avgpool{} where we average pool the model activations every two tokens and use that for the prediction stage. For presentation purposes, we illustrate the pooled values as extra tokens. 
    \textbf{e)} \offsetgist{}, a variant of \gist{} where the compressed activations at the gist positions are shifted one layer down prior to prediction to make the compressed activations immediately available to the next layer.
    \textbf{f)} \sepoffsetgist{}, which is equivalent to \offsetgist{}, except  the compressed activations are computed with a \textcolor{MainFigGreen}{separate set of model parameters}.
    \textbf{g)} \newgist{}, our proposed in-context compression method.
    The key features of \newgist{} are: (1) shifting the activations down by one layer during the prediction phase; (2) \textcolor{MainFigGreen}{compression-specific parameters} are separated from the other model parameters; (3) spreading out the tokens uniformly across the context and modifying the mask. For illustration purposes, a mask attending to the previous two pooling windows is shown but the experiments are performed with a mask attending to the previous 5 windows (see \Cref{sec:spreading_the_tokens} for details).
    }
    \label{fig:all_methods}
\end{figure}

\section{Preliminaries}
\label{sec:background}

\paragraph{In-context compression.}

We aim to compress large amounts of contextual information for downstream uses.
Consider a sequence of context tokens $T^c$.
Take $\Xc$ to be the KV-cache of $T^c$.
Our goal is to compress this context into a significantly smaller representation $\tilde\Xc$, while still generating accurate answers $T^a$ to various queries $T^q$ related to $T^c$.
The compressed representation $\tilde{\Xc}$ should have a significantly smaller size than $\Xc$, with a compression ratio $\xi$, such that $|\tilde{\Xc}| \approx |\Xc|/\xi$. Once $\tilde{\Xc}$ is computed, the original context $\Xc$ can be discarded for more efficient processing of queries.
The setup has two phases:
\begin{enumerate}
\item \emph{Compression phase}: the compressed representation $\tilde{\Xc}$ is generated from the activations $\Xc$ of context tokens $T^c$. The original context is discarded.
\item \emph{Prediction phase}: the compressed representation $\tilde{\Xc}$ and the query $T^q$ are used to generate the answer $T^a$.
\end{enumerate}

The context and query are always available at both training and inference time but the answer is only available during training and is autoregressively generated at inference time.

\begin{table}[h]
\addtolength{\tabcolsep}{-0.45em}
\caption{\textbf{Datasets in the paper.} Context length is the part of the sample that is compressed, i.e., the instruction for \alpaca{} or the story/background information for the other datasets. Sample length refers to the total token count including context, query and answer. Lengths are measured in tokens for the \gemma{} tokenizer.}
\label{tab:datasets}
\centering
\scriptsize
\begin{tabular}{@{}llrrrr@{}}
\toprule
Dataset &
  Type &
  \multicolumn{1}{l}{\begin{tabular}[c]{@{}r@{}}Avg. context\\ length\end{tabular}} &
  \multicolumn{1}{l}{\begin{tabular}[c]{@{}r@{}}Avg. sample\\ length\end{tabular}} &
  \multicolumn{1}{l}{\begin{tabular}[c]{@{}r@{}}Training\\ samples\end{tabular}} &
  \multicolumn{1}{l}{\begin{tabular}[c]{@{}r@{}}Test\\samples \end{tabular}} \\ \midrule
\alpaca{}     & Instructions & 20    & 70    & 133,441 & 1,000  \\
\squad{}       & Q\&A                  & 163   & 185   & 87,559  & 10,570 \\
\drop{}        & Q\&A                  & 305   & 327   & 77,399  & 9,536  \\
\race{}        & Q\&A/Mult. choice & 350   & 415   & 87,863  & 4,887  \\
\narrative{} & Q\&A                  & 732   & 755   & 65,494  & 6,922  \\
\fairytale{} & Q\&A                  & 3,475 & 3,501 & 8,548   & 1,025  \\ \bottomrule
\end{tabular}
\end{table}

\paragraph{Gisting.}
\gist{}~\citep{mu2024learning} is a simple in-context compression technique that leverages special gist tokens $T^g=\{g_1,\ldots,g_n\}$, which are inserted between the context $T^c$ and the query $T^q$.
The attention mask is adjusted to prevent query and answer tokens from directly attending to the context, forcing them to rely solely on the information captured by the gist activations, creating an information bottleneck\footnote{We also ensure that the mask allows all tokens to attend to the beginning-of-sequence (BOS) token, as models tend to rely on it as an attention sink~\citep{xiao2024efficient}.} (see \Cref{fig:all_methods}c).
Gist tokens have unique token ids, their own input embeddings and, importantly, do not depend on the context $\Xc$. During the compression phase, the gist activations $\Xg$ are computed given the context $\Xc$ using the model and become the compressed representation $\tilde{\Xc}$.
All model parameters are fine-tuned (including the input embeddings of the gist tokens) with perplexity loss over the answers.
Other loss functions, such as KL divergence to the base model logits, can also be used.
While \citet{mu2024learning} used a fixed number of gist tokens, leading to a variable compression rate $\xi$, we propose fixing $\xi$ and adjusting the number of gist tokens.
This is better-suited for the wide range of lengths in real-world contexts.

\paragraph{Datasets.} 

To study compression across context lengths, we selected six datasets with varying lengths.
These include the instruction-following \alpaca{}, combining Self-Instruct~\citep{wang2023selfinstruct} and Stanford Alpaca~\citep{alpaca}, 20-token average instruction length, sometimes missing queries, included for alignment with \citet{mu2024learning}.
We also consider Q\&A datasets: \squad{} (v1.1.0, Wikipedia paragraphs with crowd-sourced questions, \citealt{rajpurkar2016squad}), \drop{} (finding and processing references in longer Wikipedia paragraphs, \citealt{dua2019drop}), \race{} (English exams in China, multiple-choice, \citealt{lai2017large}), \narrative{} (book/movie summaries, \citealt{kocisky2018narrativeqa}) and \fairytale{} (expert-written questions for fairytales, \citealt{xu2022fantastic}).
We focus on Q\&A due to their balance of large context length and easily assessed responses.
Context lengths range from 163 (\squad{}) to 3,475 (\fairytale{}) tokens (see \Cref{tab:datasets}).
Dataset samples can be found in \Cref{app:dataset_samples}.

\paragraph{Evaluation.} We evaluate the perplexity loss over the answer tokens for the evaluation split of the corresponding dataset.
However, the perplexity loss is not a direct predictor of task performance.
As all datasets except \race{} have open answers, automated metrics may not fully capture the quality of the results.
Instead, we propose to use \geminijudge{} evaluation, where we use Gemini 1.5 Flash to judge the quality of the answer given the full uncompressed context (see the prompt that we used  in \Cref{app:gemini_judge}).
We report both the loss and the fraction of answers the judge considered incorrect (Gemini Score).
For both metrics, lower values indicate better performance.

\paragraph{Performance baselines.}
It is not immediately obvious what the target performance for compression should be.
It is unlikely that the compressed performance would surpass that of the base model.
Therefore, our upper baseline is fine-tuning the base model with access to the whole context (\fullbaseline{}, \Cref{fig:all_methods}a).
On the other hand, for the lower baseline we consider the model without access to any context information (\nobaseline{}, \Cref{fig:all_methods}b).
This lower baseline is not zero accuracy, as some queries are answerable through common sense.
Both baselines are trained with perplexity loss on the answer tokens.

\paragraph{\geminicompress{} baseline.} The baselines above primarily serve as performance boundaries (upper and lower bounds).
A more realistic baseline is to compress the text to a specified compression rate by prompting an LLM.
We then provide the compressed context to the same LLM and evaluate its performance using \geminijudge{}.
Since this does not involve model training, it does not produce an evaluation loss.
However, the \geminijudge{} score provides a valuable comparison point.
Appendix \ref{app:gemini_compress} contains examples of the prompts we used for compression.

\begin{figure*}
    \centering
    \includegraphics[width=0.8\textwidth]{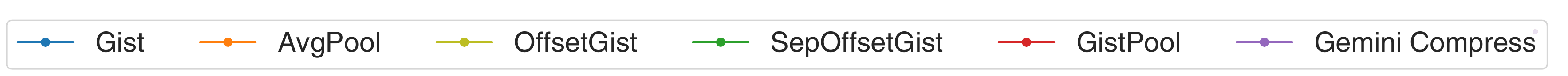}
    \includegraphics[width=1\textwidth]{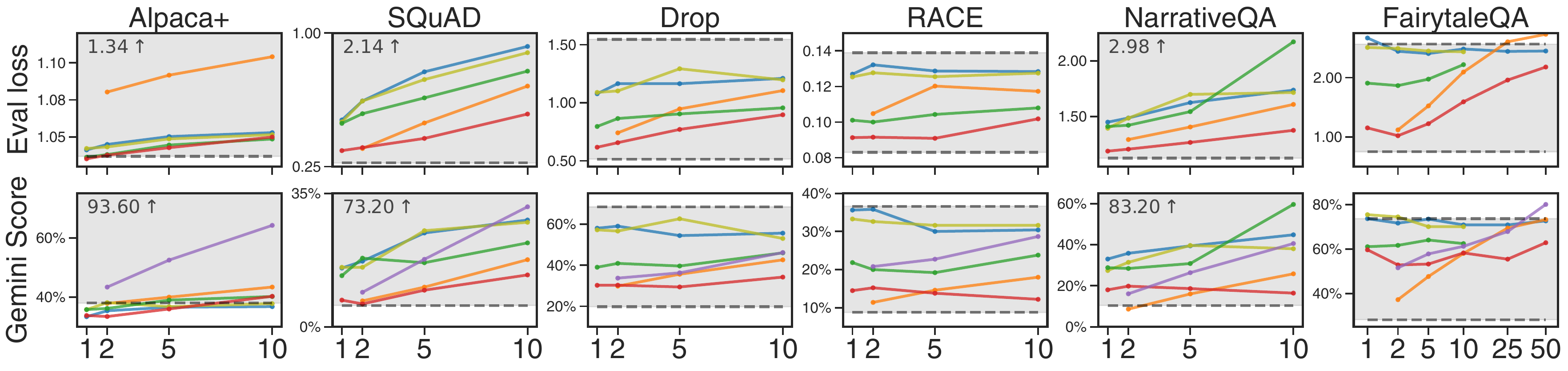}
    \makebox[\textwidth]{\raisebox{10pt}{\small \fontfamily{DejaVuSans-TLF}\selectfont Compression rate}}
    \vspace{-25pt}
    \caption{\textbf{Final evaluation loss and Gemini Score (number of errors as determined by \geminijudge{}).} \fullbaseline{} (lower) and \nobaseline{} (higher) are indicated with dashed black lines (shaded between). If the \nobaseline{} is significantly worse than other methods, we omit it for clarity and instead show its value. Lower values are better for both metrics.}
    \label{fig:eval}
\end{figure*}

\section{Gisting Fails to Compress Long Contexts}
\label{sec:gisting_fails}

\textbf{Gisting works well for compressing short instructions.}
\gist{} is fully compatible with the decoder-only transformer architecture, meeting our \archiconst{} requirement, making it a great candidate for a scalable in-context compression solution.
Thus, we evaluated \gist{} on the datasets described in \cref{sec:background} at $1\times$, $2\times$, $5\times$ and $10\times$ compression rates, plus $25\times$ and $50\times$ for \fairytale{}.
We use \gemma{} 2B \citep{gemma2} as our base model\footnote{Every other layer of \gemma{} has local attention with sliding window of 4096. Hence, when the gist tokens are interspersed, early ones might be masked out. This affects only \fairytale{} and can explain why for it the \newgist{} performance at low compression rates is lower than what one would expect.}, adding only new gist token embeddings and adjusting the attention mask as described in \Cref{sec:background}.
We fine-tuned on \race{} for one epoch, \alpaca, \squad, \narrative, and \fairytale{} for two epochs, and \drop{} for three epochs.
We used Adafactor~\citep{shazeer2018adafactor} and a constant learning rate of \num{1e-7}.

For \alpaca{}, consistent with prior work~\citep{mu2024learning}, \gist{} maintains performance up to $10\times$ compression. Accuracy remains near the \fullbaseline{} (\Cref{fig:eval}), with a similarly negligible change in the loss. This represents a virtually insignificant performance reduction and aligns with the findings reported by~\citet{mu2024learning}.

\textbf{\gist's effectiveness diminishes with longer contexts.}
While \alpaca{} focuses on short instructions (with an average instruction length around 20 tokens), other datasets feature significantly longer contexts (163--3,475 tokens) and more diverse content. 
As seen in \Cref{fig:eval}, errors increase significantly across these datasets at $10\times$ compression, failing the \scalability{} requirement.
Still, performance generally remains above the \nobaseline{} and is comparable to the \geminicompress{} baseline, suggesting some information compression.
This scalability limitation, however, is impractical for most applications.

\paragraph{\gist{} has performance issues even in the $\bm{1\times}$ compression case.}
Surprisingly, \gist{} struggles to ``compress'' longer datasets even in the $1\times$ case (i.e.\ with as many gist tokens as context tokens), with performance being noticeably worse than \fullbaseline{} for all datasets, except \alpaca{}.
In fact, for \race{} and \fairytale{} the $1\times$ \gist{} performance is close to the \nobaseline{}.
In principle, $1\times$ compression should be near lossless, as the model just needs to copy the context embeddings to the corresponding gist token positions.
Therefore, \gist{} also fails to have \losslessT{}, indicating a likely systematic problem with the original formulation of \gist{}.
In \Cref{sec:improved_gisting} we analyze what hinders \gist's effectiveness with longer sequences and propose fixes that improve its performance.

\section{Average Pooling is Unreasonably Effective}
\label{sec:average_pool}
Inspired by the observation that \gist{} lacks the \losslessT{} property, we explore a simple compression baseline that transitions to lossless compression at low rates.
Copying the context is a straightforward way to achieve lossless compression at a $1\times$ rate.
We can extend this to $\xi\times$ compression by simply averaging every $\xi$ context activations into a single compressed activation (\Cref{fig:all_methods}d).\footnote{We also add a non-compressed BOS token at the beginning of the compressed sequence.}
We call this \avgpool{}.
While we still fine-tune the model, the compression phase is parameter-free, making it a lightweight method.
\avgpool{} is easy to add to existing implementations, meeting the \archiconst{} requirement. 

\Cref{fig:eval} shows that \avgpool{} reaches significantly lower loss than the \gist{} and \geminicompress{} baselines across all datasets, except for \fairytale{}\footnote{We omit the $1\times$ compression results, as they are equivalent to the full context baseline. While implementations might introduce minor variations in positional encoding, we did not observe meaningful differences between the baseline and the $1\times$ case.}. 
Furthermore, for all but the \fairytale{} dataset, \avgpool{} is close to the full-context baseline, especially at lower compression ratios $\xi$, demonstrating its \scalability{} and \losslessT{}.
\geminijudge{} assessment also shows superior downstream performance relative to \gist{}: for example, \avgpool{} has roughly half the error rate of \gist{} for \squad, \race{}, and \narrative{}.
Beyond \losslessT{}, the performance of \avgpool{} degrades more gracefully than \gist{}. 
The only exception is \alpaca{},
potentially due to its short sequences (less than 10 tokens) collapsing into single tokens.
Overall, the performance of such a simple non-parametric baseline is surprising, especially compared to \gist{}. 

\section{Overcoming the Limitations of Gisting}
\label{sec:improved_gisting}

\gist's compression performance rapidly deteriorates with increasing context length, significantly underperforming the simple \avgpool{} baseline.
Critically, \gist{} even fails in the $1\times$ compression case, lacking \losslessT{} and indicating an inability to perform a simple data copy.
We would like to improve the compression performance of \gist{} to match that of \avgpool{}, and ideally surpass it.

To address these deficiencies, we improve \gist{} by targeting three limitations: information flow interruption (\Cref{sec:gisting_causes_offset}), limited learning capacity (\Cref{sec:gisting_cant_copy}), and lack of inductive bias in the attention (\Cref{sec:spreading_the_tokens}).
Our proposed \newgist{} method incorporates solutions for these issues and outperforms \avgpool{} and \gist{}.

\begin{figure}
    \centering
    \includegraphics[page=2,width=\columnwidth]{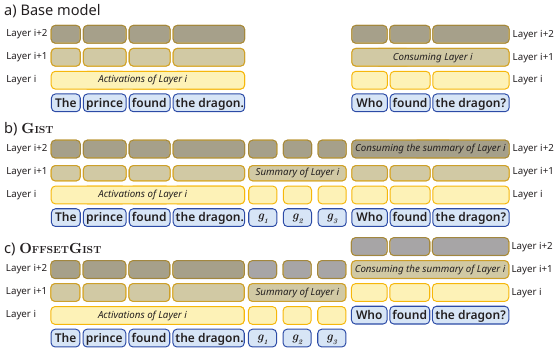}
    \vspace{-2em}
    \caption{\textbf{The gist tokens delay the information flow.} \textbf{a)} In the base model, the activations of layer $i$ are the query position inputs of layer $i+1$. \textbf{b)} The summaries introduced with \gist{} become the activations at the gist positions at layer $i+1$, which in turn become the query position inputs at layer $i+2$, one layer later than the model expects the information from layer $i$. \textbf{c)} By shifting the gist activations one layer down for the prediction stage, the summarized context from layer $i$ is available as input to the query positions at layer $i+1$ matching the expectation of the base model.}
    \label{fig:gist_flowchart}
    \vspace{-20pt}
\end{figure}

\subsection{Offsetting the activations to enable direct flow of information from context to question}
\label{sec:gisting_causes_offset}
While the \gist{} attention mask facilitates in-context compression by creating an information bottleneck, it also introduces a delay in information flow.
\Cref{fig:gist_flowchart} visually illustrates this: without the \gist{} mask, the layer $i+1$ query tokens directly attend to the layer $i$ context token outputs.
However, with the \gist{} mask, layer $i+1$ first summarizes (or copies in the $1\times$ case) layer $i$ context token outputs at the gist token positions.
This summary becomes available to query tokens only at layer $i+2$.
Consequently, layer $i$ context activations are accessible to query positions at layer $i+2$.
This delay between when the model expects context information and when it receives it potentially explains \gist{}'s failure to copy activations even in the $1\times$ case.
\gist{} can only copy previous layer outputs to the \emph{following} layer's gist positions, but for direct copying it should copy them to the \emph{current} layer's gist inputs.

There is a simple solution to this problem: move the gist activations one layer down, i.e., allow the query tokens to attend to the \emph{outputs} at the gist positions rather than the inputs.
This technique, which we call \offsetgist (see \Cref{fig:all_methods}e and 
\Cref{fig:gist_flowchart}c), aligns better with \avgpool{} by providing layer $i$ compressed activations to layer $i+1$, unlike \gist{} which provides them to layer $i+2$.

However, evaluations reveal only marginal improvements over \gist{} (see \Cref{fig:eval}).
In some cases, \offsetgist{} even slightly underperforms \gist.
Even in the $1\times$ case, \offsetgist{} exhibits significantly lower performance than the \fullbaseline{}.
Even with corrected information flow delays, the model still struggles with copying, indicating that the delay alone does not fully explain \gist{}'s poor performance at lower compression rates.

\subsection{Allowing separate parameters for summarizing}
\label{sec:gisting_cant_copy}
In principle, \offsetgist{} should enable copying context activations to the gist activations, essentially mirroring the \fullbaseline.
And yet, empirical results in \Cref{fig:eval} show that the model fails to get close to the baseline.
This performance gap suggests potentially insufficient model capacity to learn copying.
Assuming that the model activations are full-rank, lossless copying at $1\times$ compression requires structured updates to all key, value, and query matrices.
However, the model must also retain its original knowledge and skill to have high performance on the Q\&A task, limiting how much its weights can be updated.
Therefore, while the model may, in principle, be capable of copying, it might be impossible to do so without compromising performance.

To isolate the copying objective from downstream task performance, we experimented with fine-tuning the model using two separate sets of parameters: one for compressing context into gist activations, and another for processing context and query positions.
This separation permits targeted optimization for compression without impacting prediction performance.
Parameter-efficient fine-tuning methods~\citep{hu2021lora,han2024parameter} can further be employed to reduce the number of trainable parameters.
We refer to this combined approach as \sepoffsetgist{} (\Cref{fig:all_methods}f).
We also found that compression parameters require higher learning rates, hence we used $10\times$ the base rate.%

For most datasets and compression rates, we observe improvement for the \sepoffsetgist{} relative to the original \gist{} setting and \offsetgist{}, see \Cref{fig:eval}.
This reinforces our hypothesis that the model cannot perform the copying or compression and question-answering tasks at the same time. 
However, \sepoffsetgist{} with a $1\times$ compression rate still has lower accuracy than the \fullbaseline{}.
\avgpool{} (\Cref{sec:average_pool}) still performs better or comparably to \sepoffsetgist{} across the board.
Hence, even when we address these two limitations of the original \gist{} formulation ---information flow delays and learning capacity--- gisting still fails to match the performance of \avgpool{}.

\subsection{Introducing a pooling inductive bias}
\label{sec:spreading_the_tokens}

Despite the \sepoffsetgist{} model being able, at least in principle, to learn to emulate \avgpool{}, it still does not match its performance.
In \Cref{sec:attention_limitations} we offer an explanation of why the standard attention mechanism prevents \gist{} and \sepoffsetgist{} from attending to separate pooling windows, making it difficult for them to emulate \avgpool{}.
However, clearly \avgpool{} has a useful inductive bias that we would nevertheless like to use.

A simple way to ``nudge'' the model towards pooling is adjusting its attention mask: if each gist token only attends to a subset of the positions, then it would act as an attention-based generalization of \avgpool{}.
We can achieve that by interspersing the gist tokens among the context tokens, one gist token every $\xi$ context tokens, rather than placing them at the end.
Furthermore, we can restrict each gist token to only attend to a restricted set of context tokens.
In our case, we chose to attend to context tokens within the pooling windows corresponding to the previous 5 gist tokens.
Furthermore, to prevent unfairly adding additional computation for the context processing, we ensure that context tokens cannot attend to previous gist tokens (see \Cref{fig:all_methods}g).
We also ensure that all gist tokens attend to the BOS token.
Note that this is only one of many possible pooling masks.
\Cref{tab:mask_ablations} shows ablations of variations of this setup.

The query and answer tokens can still only attend to the gist positions, thus allowing discarding all context tokens for the prediction state, as in \gist{}.
Note that the same effect can be achieved by keeping the gist tokens after the context, but adjusting the attention mask to restrict what context tokens each gist token can attend to (this is the formulation we will use for the analysis in \Cref{sec:attention_limitations}).
By \newgist{} we will refer to the combination of the three techniques: shifting the gisting activations, allowing a separate set of parameters for the compression, and spacing the gist tokens uniformly along the context with the mask modified as described above.

As seen in \Cref{fig:eval}, for all models except \alpaca{}, \newgist{} outperforms \gist{} for all compression rates by a large margin.
While on par with \avgpool{} for low compression rates, it outperforms it at higher compression rates, exhibiting better \scalability{} than \avgpool{}.
\newgist{} also has \losslessT: at low compression rates, its performance is comparable to \fullbaseline.
For the $1\times$ case, it matches the \fullbaseline{} almost perfectly.
Therefore, \newgist{} satisfies all three desiderata for a scalable in-context context compression technique, while also outperforming \gist{} and \avgpool{}.

\section{Attention Without a Modified Mask Cannot Learn Average Pooling or Copying}
\label{sec:attention_limitations}

\newgist{} achieved lower loss and better downstream performance than the less constrained \sepoffsetgist{} (\Cref{sec:spreading_the_tokens}).
This raises the question: why does restricting attention improve performance?
If pooling is so good, why doesn't the model learn to do it?
We show, both experimentally and theoretically, that a transformer layer often cannot learn average pooling, even though adjusting the attention mask makes it trivial. This limitation of \gist{} justifies our modifications in \newgist{}, which effectively implements this restricted attention.
We simplify our analysis by assuming gist tokens are appended and attend only to their respective pools, which is equivalent to \newgist{} (up to the positional encodings, which we also address).

\subsection{Single layer experiments}
\label{sec:single_layer_experiments}
The \gist{} setup and its variations (\Cref{sec:improved_gisting}) summarize activations with a single attention layer.
As this attention layer should be capable of average pooling, we focus on whether a single transformer layer can perform mean pooling.

We generate a synthetic dataset with $n_\text{context}$ context tokens and $n_\text{gist}$ gist tokens, where $n_\text{context}=\xi n_\text{gist}$.
The inputs are $x = [x_1,\ldots,x_{ n_\text{context}}, g_1, \ldots, g_{n_\text{gist}} ]$, with $n_\text{context}$ embeddings sampled from the surface of the hypersphere (\sphereemb), or randomly selected embeddings from the \gemma{} vocabulary matrix (\gemmaemb), followed by $n_\text{gist}$ learnable gist embeddings.
The targets are $n_\text{context}$ zero vectors followed by $n_\text{gist}$ averaged-pooled context embeddings%
\ifshort
    .
\else
:
$$
y=
\begin{bmatrix}
0, \ldots, 0,
(\frac{1}{\xi} \sum_{i=1}^{\xi} x_i), 
\ldots,
(\frac{1}{\xi} \sum_{i=1{+}(n_\text{gist}{-}1)\xi}^{n_\text{gist}\xi} x_i)
\end{bmatrix}.
$$
\fi
We study \fixedcontext{}  for  $n_\text{context}=N$, and \varcontext{} for $n_\text{context}$ sampled from $\{N/2,N/2+\xi,\ldots,N\}$, for sequence lengths $N\in\{128,256,512\}$.
To determine if \newgist{}'s pooling bias is required, we compare two attention masks.
\Standardmask{} is the standard causal mask. \Poolmask{} restricts the $i$-th gist token to attend only to context positions from $1+(i-1)\xi$ to $i\xi$.

We employ a standard \gemma{} transformer layer with embedding size 128.
To evaluate the impact of the configuration of the attention head on copying, we test 1/1, 8/8, and the \gemma{} default 8/4 (attention heads/KV heads).
The head dimension is adjusted to maintain a constant total head dimension of 128.
Default settings are used otherwise: RMSNorm pre- and post-normalization for both attention and MLP layers, and an MLP hidden dimension of 512.
We train with the MSE loss on non-zero outputs for 50,000 steps with a batch size of 64.
To be able to compare different configurations, we report the fraction of predicted pooled embeddings which correspond to their nearest neighbour in the ground truth mean pooled embeddings.
For each configuration, we take the best performance across 3 seeds and learning rates $1\text{e-}3,1\text{e-}4,1\text{e-}5$.
We study compression rates of $1\times$ (copying), $8\times$, and $16\times$, with results in \Cref{app:single_layer_experiments_results}.

\begin{figure}
    \centering
    \includegraphics[width=\columnwidth]{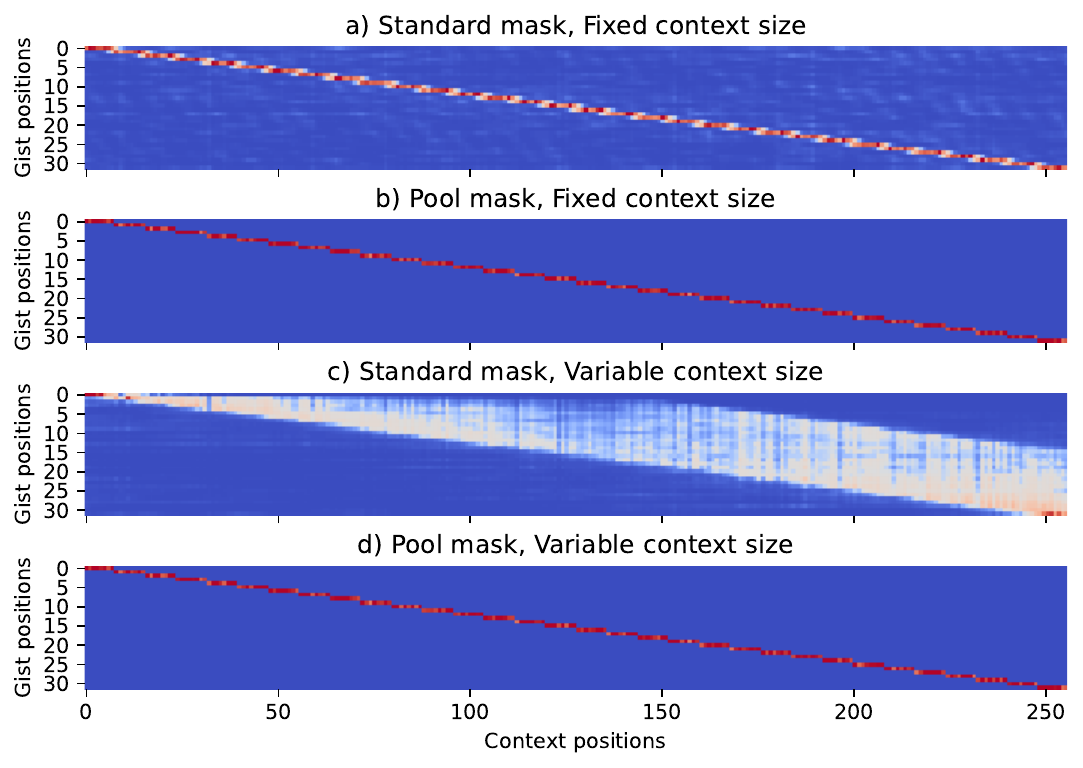}
    \vspace{-1em}
    \caption{\textbf{Attention might learn average pooling for \fixedcontext{} with \standardmask{} but requires \poolmask{} in the \varcontext{} case.} Shown are the learned attention weights for mean pooling \gemmaemb{}, context length 256 and compression rate 8. Only the attention weights for the gist positions attending to the context positions are shown. \Poolmask{} forces each gist token to attend to its corresponding group. However \standardmask{} cannot learn average pooling in the \varcontext{} case, as seen by the dispersed attention in \textbf{c)}.}
    \label{fig:learned_attention_weights}
\end{figure}

\textbf{With fixed context size, the transformer layer learns to pool \gemma{} vocabulary embeddings but not random inputs.}
In the  \fixedcontext{} with \sphereemb{} setting, the \poolmask{} has near perfect performance, while \standardmask{} performs much worse, especially for $\xi\texttt{=}1$.
However, with \gemmaemb{}, the performance is comparable, showing successful pooling.
This is reflected in the similar learned attention weights (\Cref{fig:learned_attention_weights}a,b).
This suggests that the \gemma{} vocabulary embeddings likely have learned a structure making them more amenable to pooling than general inputs, possibly due to lower-dimensionality and orthogonality to the RoPE encodings.

\textbf{When the context size varies, the transformer layer cannot learn to pool any inputs.}
Possibly, the layer learns pooling with \fixedcontext{} as the gist tokens have a fixed relative position to the context tokens.
But with \varcontext{}, these relative positions change, potentially hindering learning.
\Cref{app:single_layer_experiments_results} (\varcontext{} setup) confirms this: for both \gemmaemb{} and \sphereemb{}, across all setups, the \standardmask{} performance is significantly reduced.
This shows the transformer layer's failure to learn mean pooling with variable context unless the mask restricts attention.
Dispersed attention weights (\Cref{fig:learned_attention_weights}c) further demonstrate the inability of attention to ``focus''.
Thus, our changes in \Cref{sec:improved_gisting} are crucial for effective compression matching \avgpool{}.

\textbf{Fixing the gist embedding positions helps but does not match the pool mask performance.}
One possible explanation of the discrepancy between the \fixedcontext{} and \varcontext{} setups is that the positions of the gist tokens vary in the second, which could act as a perturbation to the gist embeddings learning precise queries for specific ranges.
To study this, we repeat the same experiment but with the positions of all gist embeddings set to 0, regardless of where they are in the sequence.
\Cref{app:single_layer_experiments_results} shows that this improves the \standardmask{} setups significantly, albeit still falling short of the \poolmask{} near-perfect performance.
This is in contrast to prior work that requires two layers for copying~\citep{olsson2022context}.
Therefore, positional encodings do play a crucial role (as investigated theoretically in the following section).
This further supports \newgist{} spreading the gist tokens across the sequence: as each gist token now has a fixed position, this is equivalent to freezing their positional embeddings.

\subsection{Fundamental limitations of the attention mechanism for copying and pooling}
\label{sec:theory}

\Cref{sec:single_layer_experiments} showed \gemma{}'s attention layer struggles with copying and attention pooling under variable context.  While comprehensive, those experiments used \gemma{}'s default architecture.
Therefore, they don't preclude other transformer layer variants (e.g., different norms/positional encodings) from learning mean pooling.
This section argues our findings are universal to the core attention mechanism, thus applying to all transformers.
See \Cref{app:constructions} for formal details.

\textbf{It is easy to construct attention mechanisms that perform copying and pooling but these require unreasonable assumptions.}
First, let's construct a transformer that copies the input at position $i$.
Using unique, unit-norm, and maximally spaced-out positional embeddings, the dot product of the positional embedding $\pi_i$ for position $i$ with itself is 1, and strictly lower for all the rest.
We can maximize the pre-softmax logits for position $i$ with suitable key and query matrices.
To approximate copying the $i$-th position with arbitrary precision we scale the query matrix with $F$, ensuring most of the attention is placed on the $i$-th position.
The full details of this construction can be found in \Cref{sec:constructions_copy_unrestricted}.
However, this approach requires arbitrarily large parameters, which would cause numerical issues and training instabilities.
Therefore, this construction is not realistic and would not be something that a model would learn.

The case with average pooling is analogous.
Positional encodings within a pooling window should be close (high dot product), and those in different windows far (low dot product).
Again, a large enough query scaling $F$ ensures the attention is approximately uniform over the pooling window and approximately zero over all other positions.
The formal analysis is in \Cref{sec:constructions_pool_unrestricted}.
Like the copying case, this relies on unbounded query scaling, impractical in reality.
It also requires some neighboring encodings being much closer than others, a property typically detrimental to learning and violated by most positional encodings used in practice.

Therefore, while we can construct transformers that do copy and perform average pooling, these are not realistic.
Let's now look at what happens if we restrict ourselves to the more practical setting of bounded inputs and parameters.

\textbf{If inputs and model parameters have bounded norms, attention cannot perform copying or pooling for arbitrarily long contexts.}
The outlined copying and pooling constructions rely on unrealistically large scaling of the query matrix.
However, how can we be sure there is no other construction that works with bounded norms for inputs and parameters?
In \Cref{sec:copy_restricted_magnitude}, we show that under three reasonable assumptions, standard attention cannot perform copying: \emph{i.}\ all positional embeddings have the same norm; \emph{ii.}\ query and key matrices have bounded norms; and \emph{iii.}\ the embedding dimension is constant and doesn't scale with the sequence length.
The intuition is that as the sequence length $l$ increases, the positional embeddings and their pre-softmax logits must get closer.
Bounded parameter and input magnitudes limit the pre-softmax logit scaling, restricting the gap between the highest and second-highest attention weights.
For sufficiently large $l$, this gap becomes arbitrarily small, dispersing attention and preventing sharp selection.
Since copying is a special case of average pooling ($\xi=1$), the latter is similarly impossible.
Thus, no attention-based model satisfying these three conditions can learn average pooling.

Both  copying and average pooling suffer from the dispersion of the \texttt{softmax} attention for long context lengths.
However, introducing masking restricting the attention to a fixed window size (akin to local attention), this problem is alleviated and both operations can be performed with parameters with bounded norms.
This is exactly what is achieved by adjusting the attention mask as we propose in \Cref{sec:spreading_the_tokens}.
Therefore, the theoretical findings here and in \Cref{app:constructions} highlight that it is indeed necessary to ensure that the transformer layer can focus on context subsequences. 

\section{Additional Experiments}

\begin{table*}[]
\centering
\small
\begin{tabular}{@{}lrrrrrrr@{}}
\toprule
\multicolumn{1}{r}{} & \multicolumn{3}{r}{Evaluation loss} & \multicolumn{1}{l}{\textbf{}} & \multicolumn{3}{r@{}}{Gemini Score (\% wrong answers)} \\ 
& \fairytale{} & \race{} & \drop{} &  & 
\fairytale{} & \race{} & \drop{}   \\  \midrule 
Full context baseline & 0.613 & 0.063 & 0.423 &  & 18.2\% & 4.4\%  & 12.2\% \\ \midrule
Gist (2x)             & 2.448 & 0.130 & 1.161 &  & 70.6\% & 33.5\% & 60.6\% \\
AvgPool  (2x)         & 1.118 & 0.105 & 0.730 &  & 36.8\% & 12.6\% & 29.6\% \\
GistPool (2x)         & 0.758 & 0.063 & 0.462 &  & 40.2\% & 10.8\% & 16.6\% \\ \midrule
Gist (10x)            & 2.466 & 0.125 & 1.204 &  & 72.8\% & 29.0\% & 57.2\% \\
AvgPool  (10x)        & 2.090 & 0.115 & 1.080 &  & 57.6\% & 18.0\% & 41.6\% \\
GistPool (10x)        & 1.037 & 0.069 & 0.645 &  & 38.2\% & 8.2\%  & 22.4\% \\ \midrule
No context baseline   & 2.415 & 0.104 & 1.599 &  & 71.8\% & 24.8\% & 65.6\% \\ \bottomrule
\end{tabular}
\caption{\textbf{Evaluation on  \gemma{} 9B.} Evaluation of \gist{}, \avgpool{} and \newgist{} for compression rates $2\times$ and $10\times$, as well as the \nobaseline{} and \fullbaseline{}. The performance improvements of \newgist{} are even more pronounced for the 9B model than for the 2B model.
The 9B \newgist{} significantly outperforms \avgpool{} at $2\times$ compression, when they tend to be on par for the 2B model (see \Cref{fig:eval,tab:all_results_loss,tab:all_results_gemini}).}
\label{tab:9B_results}
\end{table*}

\paragraph{Experiments on \gemma{} 9B.}
The main experiments in this paper were performed with \gemma{} 2B as a base model.
However, long context in-context compression would likely be even more important for larger models.
To this end, we also evaluated the three main methods of this paper, namely \gist{}, \avgpool{} and \newgist{}, on the much larger \gemma{} 9B.
We used \fairytale{}, \race{} and \drop{} for these experiments and kept everything else exactly the same as in the 2B setup.
The results are reported in \Cref{tab:9B_results}.
First, we observe that \gist{} does much more poorly with the larger 9B model than it does with the smaller 2B model: its performance is close to, sometimes even worse than, the \nobaseline{}.
\avgpool{} generally performs much better than \gist{} with its performance being better than the \nobaseline{} but still far from the \fullbaseline{}.
\newgist{}, however, outperforms \avgpool{} and \gist{} by a large margin and gets much closer to the \fullbaseline{}.
While for the 2B model, \avgpool{} was performing similarly to or better than \newgist{} at lower compression rates (with \newgist{} outperforming it at higher compression rates, see \Cref{fig:eval}), for the 9B model, \newgist{} outperforms \avgpool{} for both $2
\times$ and $5\times$.
Therefore, the benefits introduced by \newgist{} appear to be even more pronounced with larger models.

\paragraph{Single gist embedding experiments.}
One major disadvantage of \gist{} and all the variants discussed so far, including \newgist{}, is that the number of unique gist embeddings we need grows linearly with the sequence length.
When learning to compress sequences of length up to $N$ with compression rate $\xi$, we need $\lceil \nicefrac{N}{\xi} \rceil$ gist embeddings.
This causes several problems.
First, the number of additional model parameters  increases linearly in the sequence length, which could become prohibitive for longer context sizes.
Second, the gist embeddings with higher indices would be undertrained as they will be seen less often during training.
Third, a compression model cannot be used on inputs longer than the ones it has been trained with.
Moreover, it is highly unlikely that all these additional parameters are actually necessary.

To this end, we study how well \newgist{} works if it reuses the same gist embedding for all gist tokens.
The results in \Cref{tab:single_gist_variants} show that using only one gist embedding results in comparable performance across \squad{}, \drop{}, \race{}, \narrative{} and \fairytale{} for compression rates $2\times, 5\times$ and $10
\times$.
In fact, for some setups the model with a single gist embedding slightly outperforms the original \newgist{} model.
Therefore, one can make \newgist{} length-invariant by using only one gist embedding with little to no performance cost.

We also considered giving the gist tokens separate positional encodings.
This setup has two independent position counters: one for the non-gist tokens and one for the gist tokens.
\Cref{tab:single_gist_variants} no consistent improvements, so we stick with a common position counter.

\paragraph{Cross-dataset evaluation.} 
\newgist{}, and all other variants of \gist{}, being \emph{learned} methods, depend on the choice of training dataset.
That is why we also study the transferability of the trained models to other datasets.
\Cref{tab:cross_eval_gistpool} shows a matrix of \newgist{} trained on \drop{}, \fairytale{}, \narrative{}, \race{} and \squad{} for $5\times$ compression evaluated on the other four datasets.
Due to the differences in length between the datasets, we used the length-invariant \newgist{} with a single gist token embedding as described above.
In general, we see that a mismatch between the training and test datasets leads to a drop in performance.
Still, some dataset pairs fare better than others.
For example, the model trained on \squad{} has 19.8\% errors on \narrative{} while the one trained on \narrative{} has 16\%.
The limited cross-dataset performance can be attributed to the models learning dataset-specific information and output formats during the fine-tuning.
Consequently, evaluation on a different dataset introduces a significant distribution shift, leading to performance degradation.
This is likely also exacerbated by the small model size.

To study whether this drop in performance is due to the compression or the prediction stages, we perform the same cross-dataset evaluation for \avgpool{}.
As \avgpool{} does not have a learnable compression stage, this quantifies the effect of fine-tuning the predictor for a specific dataset.
The results in \Cref{tab:cross_eval_avgpool} show that \avgpool{} has lower transferability than \newgist{} across every pair of datasets.
Therefore, the lack of transferability appears to arise from the prediction, rather than the compression stage.

Additionally, we study training on a mixture of three datasets and evaluating on a fourth.
We train \newgist{} and \avgpool{} for 30,\,000 steps on $5\times$ compression with the results in \Cref{tab:loo_gistpool,tab:loo_avgpool}.
Again, there is a noticeable gap between the models trained on a mixture of other datasets and the models trained on the train split of the test dataset.
Still, \newgist{} performs consistently better than \avgpool{}.
Nevertheless, it remains an open problem how to enhance the transferability of \gist{}, \avgpool{} and \newgist{}.

\section{Related Work}

Various methods for context compression have been studied before.
Architectural modifications~\citep{beltagy2020longformer, zaheer2020big,xiao2024efficient} aim to reduce the quadratic complexity of transformers for efficient context handling.
These approaches typically require changes to the model implementation and training.
Memorizing Transformers~\citep{wu2022memorizing} explore external memory access for context compression and retrieval.
Similarly, Retrieval-Augmented Generation (RAG,~\citealp{lewis2020retrieval}) uses external knowledge. 
However, these methods add external components and, hence, system complexity.
Unlike them and other architecturally divergent approaches like T5-based encoder-decoders~\citep{li2024say} and LoCoCo's convolution-based memory~\citep{cai2024lococo}, we focus on compressing the KV-cache within the standard decoder-only transformer architecture, avoiding these complexities.

Our work, particularly the \gist{} and \newgist{} approaches, can be viewed as exploring practical implementations of the Information Bottleneck principle~\citep{tishby2000information} within the transformer architecture, aiming to learn compressed representations that retain maximal information about relevant target variables, in this case, the answers to queries.

Methods for compressing into natural language include selective token dropping~\citep{jiang2023llmlingua,li2023compressing,jiang2023longllmlingua, pan2024llmlingua,xu2024recomp, jung2024discrete} and concise summarization~\citep{wingate2022prompt, yang2023prca}. While offering interpretability and cross-model generalization, these methods can struggle with token selection and may not fully capture nuanced context.
Other methods learn embedding vectors, e.g., the In-Context Autoencoder~\citep{ge2023context} uses a few learned tokens at the end of the context.
Similar to us and~\citet{kim2024compressed}, they also use a separate set of tuned parameters for context summarization. SelfCP~\citep{gao2024selfcp} distributes these tokens throughout the sequence.
\citep{deng2024silver} also spread the gist tokens but they do not offset the activations or change the attention mask.
\citep{pang2024anchor} do change the attention mask but they also do not offset the activations.
A hierarchical compression approach is used in~\citep{chevalier2023adapting}.
Average/maximum pooling can be considered a form of token dropping/consolidation for compressing activations.
PRCA~\citep{yang2023prca} and QGC~\cite{cao2024retaining} employ query-dependent compression, potentially achieving higher efficiency.  However, our query-agnostic approach offers broader applicability and pre-computation benefits, advantageous in scenarios combining few persistent contexts with multiple diverse queries.

Average pooling is widely used in signal processing, for NLP feature aggregation and for encoding features of permutation-invariant sets~\citep{edwards2016towards, zaheer2017deep, lee2019set}. Still, its application for in-context compression is, to our knowledge, novel. Surprisingly, it outperforms \gist{} with longer contexts, a key contribution of this work. The simplicity and effectiveness of this approach underscore the need to revisit fundamental techniques when addressing long-context challenges.

\section{Discussion and conclusions}

In this paper, we address the growing challenge of large context sizes and computational costs in LLM deployment.
We found \gist{}~\citep{mu2024learning}, a method requiring only an attention mask modification, to be easily integrable with existing frameworks and systems.
However, \gist{}'s performance degrades rapidly when compressing longer contexts.
Surprisingly, average pooling performs significantly better than the learned \gist{}, despite averaging being assumed to over-smoothen and destroy information.
The effectiveness of \avgpool{} thus leaves an open question for both empirical and theoretical investigation.
Even more surprisingly, \gist{} fails to learn to emulate \avgpool{}, a phenomenon we studied extensively, both experimentally and theoretically. The success of \avgpool{} also aligns with recent research challenging the necessity of complex mechanisms for in-context learning, suggesting simpler approaches can be remarkably powerful~\citep{akyurek2022learning, von2023transformers,petrov2024prompting}.

We find that standard attention mechanisms cannot focus on single token positions or token ranges as input sequence length increases.
This aligns with similar findings by \citet{velivckovic2024softmax} that \texttt{softmax} cannot make sharp decisions.
Consequently, since the model cannot learn to focus its attention on the token groups it should summarize, we introduce constraints by modifying the attention mask.
Our experiments demonstrate that this is necessary for achieving copying and compression.
This mask modification, combined with shifting the compressed activations one layer down (to maintain correct information flow) and allowing separate compression parameters, results in \newgist{}, our proposed method that significantly outperforms \gist{}.
However, a key limitation of \newgist{} is the effective doubling of model size.
Still, we anticipate this will not be a major issue in production, as compression can occur on separate devices.
The number of parameters for the summarization model can likely be also reduced substantially using PEFT techniques like LoRA~\citep{hu2021lora}.
For scenarios with strict compute and memory constraints, \avgpool{} remains a parameter-free and computationally inexpensive alternative, often significantly outperforming \gist{}.

\section*{Impact Statement}
The goal of the work presented here is to advance the field of Machine Learning, specifically focusing on improving the efficiency of processing long context in LLMs.
There are many potential societal consequences of our work.
Improved efficiency in LLMs could lead to wider accessibility due to reduced computational costs, potentially democratizing access.
Furthermore, more efficient LLMs could have positive environmental impacts by reducing the energy consumption required for inference.
While the development of more efficient LLMs presents opportunities for positive impact, it also necessitates careful consideration of potential challenges.
This work focuses on a technical improvement, and the broader impact of more efficient LLMs requires ongoing research and discussion within the community.

\section*{Acknowledgements}
The authors would like to thank Chen Sun, Stephanie Chan, Mike Mozer and Razvan Pascanu for their feedback and support with this project.
AP acknowledges support by the EPSRC Centre for Doctoral Training in Autonomous Intelligent Machines and Systems (EP/S024050/1).

\bibliography{papers}
\bibliographystyle{acl_natbib}

\newpage
\appendix
\onecolumn

\newcommand{\mvec}[1]{{\boldsymbol{#1}}}
\newcommand{\pd}[2]{\frac{\partial {#1}}{\partial {#2}}}
\def\teq{\triangleq}
\def\vo{\mvec{o}}
\def\vz{\mvec{z}}
\def\vZ{\mvec{Z}}
\def\vq{\mvec{q}}
\def\vk{\mvec{k}}
\def\vr{\mvec{r}}
\def\vv{\mvec{v}}
\def\bvo{\bar{\mvec{o}}}
\def\bvz{\bar{\mvec{z}}}
\def\bvq{\bar{\mvec{q}}}
\def\bvk{\bar{\mvec{k}}}
\def\bvp{\bar{\mvec{p}}}
\def\bvv{\bar{\mvec{v}}}
\def\bvZ{\bar{\mvec{Z}}}
\def\mQ{{\rm{Q}}}
\def\mK{{\rm{K}}}
\def\mV{{\rm{V}}}
\def\sa{{\rm SA}}
\newcommand{\avr}[1]{\langle #1 \rangle}
\newcommand{\Avr}[1]{\left\langle #1 \right\rangle}
\newcommand{\region}[1]{{\Delta_{#1}}}
\def\balpha{\bar{\alpha}}
\def\bbeta{\bar{\beta}}
\def\bxi{\bar{\xi}}
\def\hvz{\hat{\vz}}

\def\R{\mathbb{R}}
\def\Sd{\mathbb{S}^{d-1}}

\section{Theoretical Characterization of Copying and Mean Pooling with Attention}
\label{app:constructions}

A key leitmotif in the present paper if whether a transformer layer can do copying of inputs at some positions as outputs at later positions, and whether it can do average pooling in the same fashion.
The empirical evidence in \Cref{sec:attention_limitations} showed that a \gemma{} layer can learn mean pooling in some specific cases.
However, the question of whether that property is specific to \gemma{} layers or is universal to the attention mechanism remains.
In this appendix we aim to address this via a theoretical analysis of the standard attention mechanism setup.

We first start by showing that, in general, it is relatively straightforward to construct models that perform copying or average pooling.
However, these constructions rely on some rather unrealistic assumptions, namely that the query activations can have arbitrarily large magnitude and that the positional embeddings are non-regular, i.e., some neighbouring pairs are significantly closer than other neighbouring pairs.
When we limit ourselves to the setting of bounded activations and regular positional embeddings, however, these results break.
In fact, we prove that in this more realistic setting, the attention cannot select one token or a group of consecutive tokens if the sequence length is larger than the positional embedding dimension.

\subsection{Attention can copy in the unrestricted setting}
\label{sec:constructions_copy_unrestricted}

Let's first focus on the copying case.
A simple trick to distinguish the positional information and the values that we are selecting or pooling is by separating the embedding space.
Let $\vz\in \R^{2d}$ be an embedding space.
The embedding space can then be partitioned into two subspaces $(\vz_p,\vz_v)$ with $\vz_p\in \R^d$ being a positional encoding and $\vz_v \in \R^d$ carrying the token value information.
The value matrix can be chosen in such a way that it ignores the positional component and only transforms $\vz_v$:
\begin{equation*}
    \bm V = \begin{bmatrix}
    \bm 0_{d\times d} & \bm 0_{d \times d} \\
    \bm 0_{d\times d} & \bm I_{d}
    \end{bmatrix}.
\end{equation*}
Assuming that the sequence length is at most $\ell$, we need to design a set of positional vectors in $\R^d$ that could be used to differentiate between all $\ell$ possible locations.
Consider a sphere $\Sd$ and let $\{\bm \pi_i \in \Sd\}_{i=1}^{\ell}$ be a set vectors regularly spaced on it in a sense that $\|\bm \pi_i - \bm \pi_j\| \ge \epsilon$ for all $i\ne j$ and some $0<\epsilon<2$, which depends on $l$ via the maximum spacing that can be achieved for that number of points.
Choosing these vectors $\{\bm \pi_i\}$ as positional encoding keys, we can attend to a particular location in the sequence by generating a query $\vq$ equal to $F \bm\pi_l$ of a token with a desired position $l$ and some sufficiently large $F>0$.
This can be achieved with the following query matrix:
\begin{equation*}
    \bm Q = \begin{bmatrix}
    \bm 0_{d\times d} & F \bm I_{d} \\
    \bm 0_{d \times d} & \bm 0_{d \times d}
    \end{bmatrix},
\end{equation*}
and an input structured such that $\bm x(i) = [ \ldots, \bm\pi_i] \in \R^{2d}$,
that is, the value part of the input being set equal to the positional embedding of the token we wish to select.
In that setting, with with $\bm K=\bm I_{2d\times2d}$, it is easy to see that we have
\begin{align*}
    \bm\pi_j^\top \bm K^\top \bm Q \bm x(i) &= F && \text{ for $i=j$, and} \\
    \bm \pi_j^\top \bm K^\top \bm Q \bm x(i) &\le F - \frac{\epsilon^2}{2} F && \text{ for $i\neq j$}
\end{align*}
since
\begin{gather*}
    \| \bm \pi_i - \bm\pi_j \|^2 = 2 - 2 \bm\pi_i \cdot \bm\pi_j \ge \epsilon^2.
\end{gather*}
If $F$ is chosen to be large, then the softmax operation of the self-attention with such a choice of $\bm K, \bm Q$ and $\bm V$ matrices and an input structured as $\bm x(i)$ at the $i$-th position will output a close approximation of the value at the $i$-th position.
Note that the larger $l$ is, the larger $F$ must be, both because $\epsilon$ would be smaller (more points on the sphere means they must be closer to one another) and because the softmax will be more spread-out. 

While this construct could be viable in principle, learning and maintaining such an encoding can be difficult in practice due to numerical instabilities that may arise due to attention multipliers $F \gg 2/\epsilon^2 \gg 1$ becoming large for small dimensions $d$ and a large set of possible locations $\ell$.
Moreover, contemporary architectures often employ various forms of normalization that would not allow for arbitrarily scaling of the query activations, meaning that $F$ is bounded in practice.

\subsection{Attention can perform average pooling in the unrestricted setting}
\label{sec:constructions_pool_unrestricted}

We saw in \Cref{sec:constructions_copy_unrestricted} that the attention mechanism can copy an input at a particular location by careful crafting of the positional embeddings and the $\bm K,\bm Q$ and $\bm V$ matrices, and with allowing the magnitude of $\bm Q$ to grow arbitrarily large.
The natural follow-up question is, can we select a group of inputs instead of a single input.

The setting where we want to pool all the input elements is trivial to construct.
A single self-attention layer can perform mean averaging over the entire preceding context simply by choosing $\bm Q= \bm 0_{d\times d}$ and $\bm V={\bm I_{d\times d}}$.
However, what about average pooling across a subset of tokens within the context?

It turns out that the construction in \Cref{sec:constructions_copy_unrestricted} can be generalized to support average pooling from any of the consecutive windows when multiple such windows are present in context.
The idea is to maintain a condition that $\|\bm \pi_i - \bm \pi_j\| \ge \epsilon$ for all $i\ne j$ to make sure that individual locations can be distinguished from each other, but also to group $\bm \pi_i$ from the same pooling window into a sufficiently small spherical cap, i.e., keeping them closely together.
Specifically, let us group positional encodings in such a way that $\|\bm \pi_i - \bm \pi_j\| \ge \varepsilon_\text{diff}$ for $i$ and $j$ belonging to different average-pooling windows, but $\|\bm \pi_i - \bm \pi_j\| \le \varepsilon_\text{same}$ for $i\ne j$ within the same window.
Taking the pooling window size to be $\xi$.
We can now see that with the $\bm K$ and $\bm Q$ matrices from \Cref{sec:constructions_copy_unrestricted} and $\bar{\bm x}(i) = \left[ \ldots, \nicefrac{1}{\xi} \sum_{j\in\text{window of i}} \bm \pi_j \right] \in \R^{2d}$ we get:
\begin{equation}
    \begin{aligned}
    \bm \pi_j^\top \bm K^\top \bm Q \bar{\bm x}(i) &\ge F - \epsilon_\text{same}^2F/2 && \text{ for $j$ in the same window as $i$, and} \\
    \bm \pi_j^\top \bm K^\top \bm Q \bar{\bm x}(i) &\le F - \epsilon_\text{diff}^2F/2 && \text{ for $j$ in a different window from $i$.}
    \end{aligned}
    \label{eq:result_avg_pool_large_F}
\end{equation}
Therefore, picking $F$ and $\epsilon_\text{same}$ in such a way that
\begin{gather}
\frac{2}{\varepsilon^2_\text{diff}} \ll F \ll \frac{2}{\varepsilon^2_\text{same}},
\label{eq:F_bound_avg_pool}
\end{gather}
the softmax will result in near-uniform attention over the pooling window and near-zero attention over the rest of the positions.
Combining this with the appropriate value matrix
\begin{equation*}
    \bm V = \frac{1}{\xi}\begin{bmatrix}
    \bm 0_{d\times d} & \bm 0_{d \times d} \\
    \bm 0_{d\times d} & \bm I_{d}
    \end{bmatrix},
\end{equation*}
gives us a self-attention layer that performs average pooling over a particular window, but nearly completely ignores the other tokens.

Note, however, that this construction relies on knowing the pooling groups in advance (so that we can assign them appropriate positional embeddings).
Thus the compression rate $\xi$ must be determined before the model is even trained, which is impracitcal.
Furthermore, this construction also requires that the positional embeddings are non-regular, i.e., some positional embeddings are much closer than others.
For real-world applications this is undesirable and positional embeddings used in practice tend to be regular.
As a result, $\epsilon_\text{diff}$ would be equal to $\epsilon_\text{same}$ and \Cref{eq:F_bound_avg_pool} cannot be satisfied and therefore \Cref{eq:result_avg_pool_large_F} would show similar activations for elements in the same pool group and in different pool groups.
Therefore, similarly to the construction in \Cref{sec:constructions_copy_unrestricted}, a real-world transformer is unlikely to learn to perform average pooling in the way outlined above.

\subsection{Attention cannot copy if the model weights and inputs are bounded}
\label{sec:copy_restricted_magnitude}

As discussed in \Cref{sec:constructions_copy_unrestricted}, the construction we had to enable copying relied on being able to scale the query matrix arbitrarily using the scalar factor $F$.
We commented that in practice that is not a realistic setting due to the regularization and normalization operations typically employed in neural network architectures in order to stabilize the training dynamics.
While it clear that our construction in \Cref{sec:constructions_copy_unrestricted} suffers from that limitation, it is not obvious that \emph{any} construction would have the same issue.

To this end, in this section we aim to show that, as long as one makes the (realistic assumptions) that:
\begin{enumerate}[noitemsep,topsep=0pt]
    \item The positional embeddings have the same norm;
    \item The query and key matrices have bounded norms;
    \item The embedding dimension is constant and does not scale with the sequence length,
\end{enumerate}
then, for any transformer, there exists a sufficiently long sequence such that the transformer will attribute a similar amount of attention to the sequence element it aims to copy and another sequence element that it should not copy.
Thus, making it impossible to select a single element of the sequence and perform copying.

First, consider the case where the embedding dimension is $d$ and the inputs consist solely of the $d$-dimensional positional encodings $\bm \pi_i$.
In this way, without a loss of generality, we can ignore the interactions between the values and the positional encodings.
This setting is more general than the setting where the embeddings also contain a value component, hence any impossibility result that we show in the present setting, will also hold for all the cases with values.

Hence, the problem reduces to selecting positional encodings $\bm \pi_1,\ldots,\bm \pi_d$, key and query matrices $K,Q$ and an input $\bm x$ such that
\begin{equation}
\bm \pi_i^\top \bm K^\top \bm Q \bm x > \max_{\substack{j\in[1,\ldots,l]\\j\neq i}}\bm \pi_j^\top \bm K^\top \bm Q \bm x + \Delta,
\label{eq:condition_for_copying}
\end{equation}
for some fixed $\Delta$ that depends on the difference we want between the logits of the top and the second highest predictions, in order to attend almost exclusively to the top prediction.
We claim that if $\bm \pi_i$ all lie on the sphere $\Sd$, and $\bm K, \bm Q$ and $\bm x$ have bounded $L_2$ norms, then for any fixed $\Delta$, there exists a sequence length $l$ for which \Cref{eq:condition_for_copying} must be violated.

Fix a position $i$ that we wish to copy.
Intuitively, increasing the number of points on the sphere means that the upper bound on the minimum distance between any pair of points must decrease (with the rate of decrease being related to the spherical caps packing problem, \citealp{rankin1955closest}).
Therefore, we can always find an $l$ such that
\begin{equation}
    \min_{\substack{j\in[1,\ldots,l]\\j\neq i}} \| \bm \pi_i-\bm \pi_j \| = \delta,
    \label{eq:copy_points_min}
\end{equation}
is made as small as we want.
Take $k$ to be a position that achieves the minimum in \Cref{eq:copy_points_min}, i.e., the closest positional embedding to $\bm \pi_i$.
Thus we have:
\begin{align*}
    \bm \pi_k^\top \bm K^\top \bm Q \bm x &= \bm \pi_i^\top \bm K^\top \bm Q \bm x + (\bm\pi_k-\bm\pi_i)^\top \bm K^\top \bm Q \bm x \\
    &\geq \bm\pi_i^\top \bm K^\top \bm Q \bm x - \delta \sqrt{d}\  \bm 1^\top \bm K^\top \bm Q \bm x \\
    &\geq \bm \pi_i^\top \bm K^\top \bm Q \bm x - \delta d \|\bm K\|_2 \|\bm Q\|_2 \|\bm x\|. 
\end{align*}
Now, if we select $l$ to be large enough such as $\delta < \frac{\Delta}{d \|\bm K\|_2 \|\bm Q\|_2 \|\bm x\|}$,
then 
$$
\max_{\substack{j\in[1,\ldots,l]\\j\neq i}}\bm\pi_j^\top \bm K^\top \bm Q \bm x + \Delta \geq \bm \pi_k^\top \bm K^\top \bm Q \bm x + \Delta > \bm\pi_i^\top \bm K^\top \bm Q \bm x.
$$
Hence, \Cref{eq:condition_for_copying} must be violated.
Note that we did not require any structure for the positional encodings, the $\bm K$ and $\bm Q$ matrices and $\bm x$ beyond the three assumptions above.
Therefore, this result is general and holds for most, if not all, models used in practice.

A different way to look at the same problem can be found as Lemma 2.1 in \citep{velivckovic2024softmax}.

\subsection{Attention cannot perform average pooling if the model weights and inputs are bounded}
This follows directly from \Cref{sec:copy_restricted_magnitude} because copying is a special case ($\xi=1$) of average pooling.
Thus, if the transformer model with bounded activations and key and query matrices cannot copy, then it also cannot perform average pooling.

\FloatBarrier
\newpage
\section{Comprehensive results}

\begin{table}[!htb]
\small
\centering
\begin{tabular}{@{}lcrrrrrr@{}}
\toprule
 &  \begin{tabular}[c]{@{}c@{}}Compression\\rate ($\xi$)\end{tabular} & \alpaca{} & \squad{} & \drop{} & \race{} & \narrative{} & \fairytale{} \\
\midrule
Full context baseline &  & 1.037 & 0.270 & 0.514 & 0.083 & 1.126 & 0.751 \\
\midrule
No context baseline &  & 1.339 & 2.139 & 1.546 & 0.139 & 2.975 & 2.563 \\
\midrule
\multirow{5}{*}{AvgPool} & 2 & 1.080 & 0.352 & 0.741 & 0.105 & 1.293 & 1.119 \\
 & 5 & 1.092 & 0.495 & 0.947 & 0.120 & 1.406 & 1.523 \\
 & 10 & 1.104 & 0.702 & 1.105 & 0.117 & 1.609 & 2.094 \\
 & 25 &  &  &  &  &  & 2.604 \\
 & 50 &  &  &  &  &  & 2.730 \\
\midrule
\multirow{6}{*}{Gist} & 1 & 1.041 & 0.512 & 1.077 & 0.127 & 1.450 & 2.668 \\
 & 2 & 1.045 & 0.619 & 1.165 & 0.132 & 1.487 & 2.443 \\
 & 5 & 1.050 & 0.783 & 1.164 & 0.129 & 1.625 & 2.406 \\
 & 10 & 1.053 & 0.925 & 1.210 & 0.128 & 1.737 & 2.482 \\
 & 25 &  &  &  &  &  & 2.441 \\
 & 50 &  &  &  &  &  & 2.448 \\
\midrule
\multirow{4}{*}{OffsetGist} & 1 & 1.042 & 0.501 & 1.089 & 0.125 & 1.395 & 2.508 \\
 & 2 & 1.043 & 0.619 & 1.102 & 0.128 & 1.486 & 2.491 \\
 & 5 & 1.049 & 0.739 & 1.293 & 0.126 & 1.699 & 2.448 \\
 & 10 & 1.052 & 0.890 & 1.197 & 0.128 & 1.716 & 2.435 \\
\midrule
\multirow{4}{*}{SepGist} & 1 & 1.036 & 0.505 & 0.815 & 0.109 & 1.468 & 1.874 \\
 & 2 & 1.041 & 0.592 & 0.951 & 0.108 & 1.537 & 1.943 \\
 & 5 & 1.045 & 0.732 & 1.022 & 0.114 & 1.619 & 2.012 \\
 & 10 & 1.052 & 0.764 & 1.069 & 0.110 & 1.834 & 2.124 \\
\midrule
\multirow{4}{*}{Gist with pool mask} & 1 & 1.035 & 0.306 & 0.653 & 0.105 & 1.230 & 2.617 \\
 & 2 & 1.039 & 0.337 & 0.809 & 0.107 & 1.251 & 1.895 \\
 & 5 & 1.044 & 0.433 & 1.030 & 0.107 & 1.350 & 2.163 \\
 & 10 & 1.055 & 0.630 & 1.167 & 0.105 & 1.477 & 2.433 \\
\midrule
\multirow{4}{*}{SepGist with pool mask} & 1 & 1.036 & 0.355 & 0.599 & 0.089 & 1.188 & 1.186 \\
 & 2 & 1.041 & 0.365 & 0.667 & 0.091 & 1.205 & 1.021 \\
 & 5 & 1.044 & 0.425 & 0.758 & 0.090 & 1.265 & 1.172 \\
 & 10 & 1.049 & 0.547 & 0.910 & 0.101 & 1.372 & 1.481 \\
\midrule
\multirow{4}{*}{OffsetGist with pool mask} & 1 & 1.038 & 0.312 & 0.687 & 0.105 & 1.242 & 2.771 \\
 & 2 & 1.040 & 0.332 & 0.778 & 0.102 & 1.272 & 2.355 \\
 & 5 & 1.046 & 0.453 & 0.970 & 0.108 & 1.389 & 2.338 \\
 & 10 & 1.059 & 0.670 & 1.171 & 0.112 & 1.505 & 2.426 \\
\midrule
\multirow{4}{*}{OffsetSepGist} & 1 & 1.036 & 0.493 & 0.796 & 0.101 & 1.411 & 1.906 \\
 & 2 & 1.038 & 0.547 & 0.864 & 0.100 & 1.422 & 1.865 \\
 & 5 & 1.045 & 0.636 & 0.904 & 0.104 & 1.544 & 1.974 \\
 & 10 & 1.049 & 0.786 & 0.956 & 0.108 & 2.172 & 2.219 \\
\midrule
\multirow{6}{*}{GistPool} & 1 & 1.035 & 0.340 & 0.617 & 0.091 & 1.189 & 1.153 \\
 & 2 & 1.038 & 0.357 & 0.656 & 0.092 & 1.207 & 1.020 \\
 & 5 & 1.043 & 0.408 & 0.770 & 0.091 & 1.267 & 1.223 \\
 & 10 & 1.050 & 0.545 & 0.897 & 0.102 & 1.376 & 1.592 \\
 & 25 &  &  &  &  &  & 1.959 \\
 & 50 &  &  &  &  &  & 2.178 \\
\bottomrule
\end{tabular}
\caption{\textbf{Ablation study of the improvements to \gist}. Evaluation loss for the experiments and ablations in the main text.}
\label{tab:all_results_loss}
\end{table}

\begin{table}
\small
\centering
\begin{tabular}{@{}lcrrrrrr@{}}
\toprule
 &  \begin{tabular}[c]{@{}c@{}}Compression\\rate ($\xi$)\end{tabular} & \alpaca{} & \squad{} & \drop{} & \race{} & \narrative{} & \fairytale{} \\
\midrule
Full context baseline &  & 38.0\% & 5.6\% & 19.8\% & 8.8\% & 10.4\% & 28.2\% \\
\midrule
No context baseline &  & 93.6\% & 73.2\% & 68.4\% & 36.6\% & 83.2\% & 73.6\% \\
\midrule
\multirow{5}{*}{AvgPool} & 2 & 37.9\% & 6.8\% & 29.8\% & 11.4\% & 8.6\% & 37.2\% \\
 & 5 & 40.0\% & 10.4\% & 35.6\% & 14.6\% & 16.0\% & 47.6\% \\
 & 10 & 43.4\% & 17.6\% & 42.6\% & 18.0\% & 25.8\% & 58.0\% \\
 & 25 &  &  &  &  &  & 69.4\% \\
 & 50 &  &  &  &  &  & 73.2\% \\
\midrule
\multirow{6}{*}{Gist} & 1 & 33.4\% & 15.4\% & 58.0\% & 35.6\% & 33.0\% & 73.6\% \\
 & 2 & 35.4\% & 17.2\% & 59.0\% & 35.8\% & 35.8\% & 71.6\% \\
 & 5 & 36.6\% & 24.6\% & 54.4\% & 30.0\% & 39.4\% & 73.4\% \\
 & 10 & 36.8\% & 28.0\% & 55.6\% & 30.4\% & 44.8\% & 70.8\% \\
 & 25 &  &  &  &  &  & 70.8\% \\
 & 50 &  &  &  &  &  & 72.6\% \\
\midrule
\multirow{4}{*}{OffsetGist} & 1 & 35.8\% & 15.6\% & 57.2\% & 33.3\% & 27.4\% & 75.4\% \\
 & 2 & 38.2\% & 15.6\% & 56.6\% & 32.6\% & 31.4\% & 74.4\% \\
 & 5 & 36.8\% & 25.2\% & 62.6\% & 31.6\% & 39.6\% & 70.0\% \\
 & 10 & 37.8\% & 27.4\% & 53.0\% & 31.6\% & 38.0\% & 70.0\% \\
\midrule
\multirow{4}{*}{SepGist} & 1 & 34.6\% & 16.8\% & 43.2\% & 23.7\% & 33.6\% & 64.0\% \\
 & 2 & 34.2\% & 18.0\% & 50.2\% & 22.6\% & 33.8\% & 66.0\% \\
 & 5 & 36.2\% & 21.8\% & 47.0\% & 18.2\% & 38.8\% & 68.4\% \\
 & 10 & 39.4\% & 24.4\% & 45.8\% & 24.8\% & 46.0\% & 66.8\% \\
\midrule
\multirow{4}{*}{Gist with pool mask} & 1 & 32.6\% & 7.6\% & 34.4\% & 21.4\% & 20.0\% & 76.6\% \\
 & 2 & 35.8\% & 7.4\% & 42.0\% & 20.2\% & 21.4\% & 65.0\% \\
 & 5 & 38.5\% & 11.4\% & 43.4\% & 19.0\% & 22.2\% & 65.2\% \\
 & 10 & 37.2\% & 17.8\% & 49.2\% & 21.4\% & 24.6\% & 68.4\% \\
\midrule
\multirow{4}{*}{SepGist with pool mask} & 1 & 35.0\% & 6.4\% & 28.2\% & 16.6\% & 18.2\% & 57.8\% \\
 & 2 & 34.8\% & 6.0\% & 33.8\% & 13.6\% & 20.2\% & 53.0\% \\
 & 5 & 37.2\% & 8.0\% & 29.0\% & 11.8\% & 19.0\% & 51.0\% \\
 & 10 & 35.4\% & 15.4\% & 35.4\% & 13.2\% & 20.2\% & 55.4\% \\
\midrule
\multirow{4}{*}{OffsetGist with pool mask} & 1 & 33.8\% & 7.6\% & 37.6\% & 21.0\% & 21.8\% & 79.2\% \\
 & 2 & 33.8\% & 7.4\% & 39.1\% & 22.0\% & 22.6\% & 69.5\% \\
 & 5 & 36.6\% & 11.8\% & 43.2\% & 20.6\% & 24.2\% & 70.2\% \\
 & 10 & 37.6\% & 19.4\% & 48.8\% & 21.8\% & 26.8\% & 71.8\% \\
\midrule
\multirow{4}{*}{OffsetSepGist} & 1 & 35.8\% & 13.4\% & 39.0\% & 21.8\% & 28.8\% & 61.0\% \\
 & 2 & 36.2\% & 18.0\% & 40.9\% & 20.0\% & 28.4\% & 61.6\% \\
 & 5 & 39.0\% & 16.8\% & 39.6\% & 19.2\% & 30.8\% & 64.0\% \\
 & 10 & 40.2\% & 22.0\% & 46.0\% & 23.8\% & 59.6\% & 62.4\% \\
\midrule
\multirow{6}{*}{GistPool} & 1 & 33.8\% & 7.0\% & 30.2\% & 14.5\% & 18.0\% & 59.6\% \\
 & 2 & 33.5\% & 6.0\% & 30.3\% & 15.2\% & 19.8\% & 52.8\% \\
 & 5 & 36.0\% & 9.6\% & 29.4\% & 13.8\% & 18.6\% & 53.2\% \\
 & 10 & 40.3\% & 13.6\% & 34.2\% & 12.2\% & 16.4\% & 58.2\% \\
 & 25 &  &  &  &  &  & 55.4\% \\
 & 50 &  &  &  &  &  & 62.8\% \\
\bottomrule
\end{tabular}
\caption{\textbf{Ablation study of the improvements to \gist}. Numerical values for the percentage of wrong answers (Gemini Score) according to the \geminijudge{} for the experiments and ablations discussed in the main text.}
\label{tab:all_results_gemini}
\end{table}

\begin{table}
\centering
\small
\begin{tabular}{@{}lcrrrrrr@{}}
\toprule
 &  \begin{tabular}[c]{@{}c@{}}Compr.\\rate ($\xi$)\end{tabular} &  \squad{} & \drop{} & \race{} & \narrative{} & \fairytale{} \\
\midrule
\multirow{3}{*}{Gist} & 2 & 0.619 / 17.2\% & 1.165 / 59.0\% & 0.132 / 35.8\% & 1.487 / 35.8\% & 2.443 / 71.6\% \\
 & 5 & 0.783 / 24.6\% & 1.164 / 54.4\% & 0.129 / 30.0\% & 1.625 / 39.4\% & 2.406 / 73.4\% \\
 & 10 & 0.925 / 28.0\% & 1.210 / 55.6\% & 0.128 / 30.4\% & 1.737 / 44.8\% & 2.482 / 70.8\% \\
\midrule
\multirow{3}{*}{GistPool} & 2 & 0.357 / \ \ 6.0\% & 0.656 / 30.3\% & 0.092 / 15.2\% & 1.207 / 19.8\% & 1.020 / 52.8\% \\
 & 5 & 0.408 / \ \ 9.6\% & 0.770 / 29.4\% & 0.091 / 13.8\% & 1.267 / 18.6\% & 1.223 / 53.2\% \\
 & 10 & 0.545 / 13.6\% & 0.897 / 34.2\% & 0.102 / 12.2\% & 1.376 / 16.4\% & 1.592 / 58.2\% \\
\midrule
\multirow{3}{*}{\begin{tabular}[c]{@{}l@{}}GistPool with\\single gist embedding\end{tabular}}& 2 & 0.364 / \ \ 6.2\% & 0.657 / 25.0\% & 0.083 / \ \ 9.8\% & 1.207 / 11.8\% & 0.971 / 43.4\% \\
 & 5 & 0.396 / \ \ 9.8\% & 0.726 / 32.0\% & 0.094 / 12.8\% & 1.263 / 16.0\% & 1.185 / 43.2\% \\
 & 10 & 0.575 / 13.2\% & 0.859 / 32.6\% & 0.104 / 14.4\% & 1.364 / 17.8\% & 1.483 / 48.2\% \\
\midrule
\multirow{3}{*}{\begin{tabular}[c]{@{}l@{}}GistPool with \\single gist embedding and\\separate positional\ encodings\end{tabular}} & 2 & 0.379 / \ \  5.2\% & 0.720 / 27.8\% & 0.094 / 13.0\% & 1.255 / 11.0\% & 1.270 / 47.4\% \\
 & 5 & 0.491 / \ \ 9.8\% & 0.858 / 32.6\% & 0.099 / 13.8\% & 1.340 / 13.8\% & 1.849 / 57.0\% \\
 & 10 & 0.640 / 12.6\% & 0.912 / 35.8\% & 0.099 / 15.2\% & 1.472 / 21.8\% & 2.089 / 65.0\% \\
\bottomrule
\end{tabular}
\caption{\textbf{Variants of \newgist{} using only a single gist embedding (evaluation loss and Gemini score, i.e. \% wrong answers).} The original \gist{} and \newgist{} setups are dependent on the sequence length as one needs to learn as many gist embeddings as the longest sequence divided by the compression rate. This requires additional parameters on the order of the sequence length and prevents scaling to longer sequences than trained on. To this end, we study whether we can reuse the same gist embedding for all gist tokens. We also study adding a separate counter for the positions of the gist tokens. \newgist{} with single gist embeddings performs only slightly worse than \newgist{}. Using separate positional encodings does not seem to improve performance. Therefore, for cross-dataset evaluations (\Cref{tab:cross_eval_gistpool,tab:loo_gistpool}) we use \newgist{} with single gist embeddings.}
\label{tab:single_gist_variants}
\end{table}

\begin{table}[]
\centering
\begin{tabular}{@{}lrrrrr@{}}
\toprule
             &   \multicolumn{5}{c}{Training dataset}\\ 
Test dataset & \drop{}  & \fairytale{} & \narrative{}  & \race{}  & \squad{} \\ \midrule
\drop{} & 0.726 / 32.0\% & 1.901 / 77.6\% & 1.838 / 63.4\% & 1.900 / 73.6\% & 2.235 / 68.8\% \\
\fairytale{} & 3.558 / 64.2\% & 1.185 / 43.2\% & 2.049 / 53.2\% & 3.463 / 88.2\% & 3.028 / 75.6\% \\
\narrative{} & 3.102 / 26.2\% & 1.849 / 29.4\% & 1.263 / 16.0\% & 2.671 / 57.6\% & 2.886 / 19.8\% \\
\race{} & 0.300 / 37.6\% & 1.302 / 49.8\% & 0.374 / 30.6\% & 0.094 / 12.8\% & 0.339 / 35.4\% \\
\squad{} & 0.974 / 18.4\% & 1.620 / 50.0\% & 0.871 / 16.4\% & 2.142 / 57.8\% & 0.396 / \ \ 9.8\% \\ \bottomrule
\end{tabular}
\caption{\textbf{Cross-dataset evaluation of \newgist{} with a single gist token.} Shown are the numerical values for the evaluation loss and the percentage of wrong answers (Gemini Score) when \gemma{} 2B is trained with \newgist{} with the dataset corresponding to the column and evaluated on the dataset corresponding to the row. A single gist token was used in order to allow for transferability across datasets of different lengths. The results are for \newgist{} with $5\times$ compression.}
\label{tab:cross_eval_gistpool}
\end{table}

\begin{table}[]
\centering
\begin{tabular}{@{}lrrrrr@{}}
\toprule
             &   \multicolumn{5}{c}{Training dataset}\\ 
Test dataset & \drop{}  & \fairytale{} & \narrative{}  & \race{}  & \squad{} \\ \midrule
\drop{} & 0.947 / 35.6\% & 7.190 / 81.6\% & 1.951 / 69.8\% & 3.806 / 74.7\% & 2.884 / 72.3\% \\
\fairytale{} & 4.877 / 79.0\% & 1.523 / 47.6\% & 2.490 / 63.0\% & 9.317 / 87.6\% & 4.225 / 81.0\% \\
\narrative{} & 3.207 / 34.4\% & 5.541 / 56.6\% & 1.406 / 16.0\% & 7.758 / 71.4\% & 3.434 / 28.0\% \\
\race{} & 0.679 / 37.4\% & 8.784 / 36.3\% & 0.815 / 37.0\% & 0.120 / 14.6\% & 0.580 / 43.1\% \\
\squad{} & 1.273 / 25.6\% & 9.686 / 83.8\% & 1.445 / 30.0\% & 7.905 / 74.2\% & 0.495 / 10.4\% \\ \bottomrule
\end{tabular}
\caption{\textbf{Cross-dataset evaluation of \avgpool{}.} Shown are the numerical values for the evaluation loss and the percentage of wrong answers (Gemini Score) when \gemma{} 2B is trained with \avgpool{} with the dataset corresponding to the column and evaluated on the dataset corresponding to the row. The results are for \avgpool{} with $5\times$ compression.  The cross-dataset evaluation performance for \avgpool{} is consistently worse than that for \newgist{} in \Cref{tab:cross_eval_gistpool}.}
\label{tab:cross_eval_avgpool}
\end{table}

\clearpage

\begin{table}[]
\centering
\begin{tabular}{@{}lcrrrr@{}}
\toprule
             & \multicolumn{1}{r}{}                 & \multicolumn{2}{r}{Trained on the other 3 datasets} & \multicolumn{2}{r@{}}{Trained on the same dataset} \\ 
Test dataset & \multicolumn{1}{l}{Compression rate ($\xi$)} & Eval. loss              & Gemini Score              & Eval. loss      & Gemini Score      \\ \toprule
\multirow{3}{*}{\drop{}}  & 2  & 1.573 & 59.4\% & 0.657 & 25.0\% \\
                       & 5  & 1.581 & 63.8\% & 0.726 & 32.0\% \\
                       & 10 & 1.641 & 67.4\% & 0.859 & 32.6\% \\ \midrule
\multirow{3}{*}{\narrative{}}  & 2  & 2.069 & 12.2\% & 1.207 & 11.8\% \\
                       & 5  & 2.012 & 19.2\% & 1.263 & 16.0\% \\
                       & 10 & 2.039 & 24.6\% & 1.364 & 17.8\% \\ \midrule
\multirow{3}{*}{\race{}}  & 2  & 0.324 & 28.0\% & 0.083 & 9.8\%  \\
                       & 5  & 0.302 & 28.6\% & 0.094 & 12.8\% \\
                       & 10 & 0.277 & 31.8\% & 0.104 & 14.4\% \\ \midrule
\multirow{3}{*}{\squad{}} & 2  & 0.571 & 8.4\%  & 0.364 & 6.2\%  \\
                       & 5  & 0.725 & 15.6\% & 0.396 & 9.8\%  \\
                       & 10 & 1.007 & 18.2\% & 0.575 & 13.2\% \\ \bottomrule
\end{tabular}
\caption{\textbf{Leave-one-out cross-dataset evaluation of \newgist{} with a single gist token.} Given a test dataset, \gemma{} 2B is trained with \newgist{} on a mixture of the other three datasets and evaluated on the test dataset. Both the evaluation loss and the percentage of wrong answers (Gemini Score) are reported. For convenience, we also report the loss and Gemini Score when training and evaluating on the same dataset. The leave-one-out performance for some datasets (\narrative{} and \squad{}) is close to that of the models specifically trained for these datasets, indicating universality and transferability of the learned compression scheme. For others (\drop{} and \race{}), there are more pronounced differences in performance.}
\label{tab:loo_gistpool}
\end{table}

\begin{table}[]
\centering
\begin{tabular}{@{}lcrrrr@{}}
\toprule
             & \multicolumn{1}{r}{}                 & \multicolumn{2}{r}{Trained on the other 3 datasets} & \multicolumn{2}{r@{}}{Trained on the same dataset} \\ 
Test dataset & \multicolumn{1}{l}{Compression rate ($\xi$)} & Eval. loss              & Gemini Score              & Eval. loss      & Gemini Score      \\ \toprule
\multirow{3}{*}{\drop{}}      & 2    & 1.821 & 64.0\% & 0.741 & 29.8\% \\
                              & 5    & 1.973 & 64.6\% & 0.947 & 35.6\% \\
                              & 10   & 2.014 & 68.0\% & 1.105 & 42.6\% \\ \midrule
\multirow{3}{*}{\narrative{}} & 2    & 2.769 & 17.6\% & 1.293 & 8.6\% \\
                              & 5    & 2.630 & 25.8\% & 1.406 & 16.0\% \\
                              & 10   & 2.721 & 39.0\% & 1.609 & 25.8\% \\ \midrule
\multirow{3}{*}{\race{}}      & 2    & 0.284 & 36.4\% & 0.105 & 11.4\% \\
                              & 5    & 0.345 & 37.4\% & 0.120 & 14.6\% \\
                              & 10   & 0.364 & 41.2\% & 0.117 & 18.0\% \\ \midrule
\multirow{3}{*}{\squad{}}     & 2    & 0.674 & 9.6\%  & 0.352 & 6.8\% \\
                              & 5    & 1.127 & 20.4\% & 0.495 & 10.4\% \\
                              & 10   & 1.599 & 34.0\% & 0.702 & 17.6\% \\ \bottomrule
\end{tabular}
\caption{\textbf{Leave-one-out cross-dataset evaluation of \avgpool{}.} Given a test dataset, \gemma{} 2B is trained with \avgpool{} on a mixture of the other three datasets and evaluated on the test dataset. Both the evaluation loss and the percentage of wrong answers (Gemini Score) are reported. For convenience, we also report the loss and Gemini Score when training and evaluating on the same dataset. The leave-one-out performance for \avgpool{} is consistently worse than that for \newgist{} in \Cref{tab:loo_gistpool}.}
\label{tab:loo_avgpool}
\end{table}

\begin{table}
\centering
\small
\begin{tabular}{cccccrrr}
\toprule
\begin{tabular}[c]{@{}c@{}}Number of\\pooling\\windows\end{tabular} &
\begin{tabular}[c]{@{}c@{}}Contexts\\attend to\\gists\end{tabular} &
\begin{tabular}[c]{@{}c@{}}Gists\\ attend to\\ gists\end{tabular} &
\begin{tabular}[c]{@{}c@{}}Gists\\ attend to\\ self\end{tabular} &
\begin{tabular}[c]{@{}c@{}}Gists\\ attend to\\ BOS\end{tabular} &
\drop{} &
\fairytale{} &
\squad{} \\ \midrule
\multirow{16}{*}{1} & \multirow{8}{*}{False} & \multirow{4}{*}{False} & \multirow{2}{*}{False} & False & 0.879 & 1.472 & 0.463 \\
                           &                        &                        &                        & True  & 0.812 & 1.315 & 0.468 \\  \cmidrule(r){4-8}
                           &                        &                        & \multirow{2}{*}{True}  & False & 0.814 & 1.335 & 0.443 \\
                           &                        &                        &                        & True  & 0.808 & 1.330 & 0.449 \\ \cmidrule(r){3-8}
                           &                        & \multirow{4}{*}{True}  & \multirow{2}{*}{False} & False & 0.896 & 1.398 & 0.458 \\
                           &                        &                        &                        & True  & 0.806 & 1.287 & 0.447 \\  \cmidrule(r){4-8}
                           &                        &                        & \multirow{2}{*}{True}  & False & 0.787 & 1.295 & 0.442 \\
                           &                        &                        &                        & True  & 0.822 & 1.318 & 0.428 \\ \cmidrule(r){2-8}
                           & \multirow{8}{*}{True}  & \multirow{4}{*}{False} & \multirow{2}{*}{False} & False & 0.871 & 1.429 & 0.471 \\
                           &                        &                        &                        & True  & 0.832 & 1.307 & 0.432 \\  \cmidrule(r){4-8}
                           &                        &                        & \multirow{2}{*}{True}  & False & 0.788 & 1.317 & 0.462 \\
                           &                        &                        &                        & True  & 0.800 & 1.323 & 0.442 \\  \cmidrule(r){3-8}
                           &                        & \multirow{4}{*}{True}  & \multirow{2}{*}{False} & False & 0.807 & 1.384 & 0.450 \\
                           &                        &                        &                        & True  & 0.817 & 1.279 & 0.435 \\  \cmidrule(r){4-8}
                           &                        &                        & \multirow{2}{*}{True}  & False & 0.771 & 1.317 & 0.448 \\
                           &                        &                        &                        & True  & 0.820 & 1.275 & 0.438 \\ \midrule
\multirow{16}{*}{\textbf{5}} & \multirow{8}{*}{\textbf{False}} & \multirow{4}{*}{False} & \multirow{2}{*}{False} & False & 0.774 & 1.264 & 0.427 \\
                           &                        &                        &                        & True  & 0.793 & 1.220 & 0.432 \\  \cmidrule(r){4-8}
                           &                        &                        & \multirow{2}{*}{True}  & False & 0.789 & 1.229 & 0.394 \\
                           &                        &                        &                        & True  & 0.729 & 1.212 & 0.406 \\  \cmidrule(r){3-8}
                           &                        & \multirow{4}{*}{\textbf{True}}  & \multirow{2}{*}{False} & False & 0.782 & 1.246 & 0.413 \\
                           &                        &                        &                        & True  & 0.784 & 1.216 & 0.422 \\  \cmidrule(r){4-8}
                           &                        &                        & \multirow{2}{*}{\textbf{True}}  & False & 0.730 & 1.226 & 0.410 \\
                           &                        &                        &                        & \textbf{True}  & \textbf{0.770} & \textbf{1.223} & \textbf{0.408} \\ \cmidrule(r){2-8}
                           & \multirow{8}{*}{True}  & \multirow{4}{*}{False} & \multirow{2}{*}{False} & False & 0.759 & 1.250 & 0.404 \\
                           &                        &                        &                        & True  & 0.762 & 1.200 & 0.408 \\  \cmidrule(r){4-8}
                           &                        &                        & \multirow{2}{*}{True}  & False & 0.765 & 1.237 & 0.411 \\
                           &                        &                        &                        & True  & 0.771 & 1.274 & 0.400 \\  \cmidrule(r){3-8}
                           &                        & \multirow{4}{*}{True}  & \multirow{2}{*}{False} & False & 0.803 & 1.232 & 0.421 \\
                           &                        &                        &                        & True  & 0.763 & 1.226 & 0.412 \\  \cmidrule(r){4-8}
                           &                        &                        & \multirow{2}{*}{True}  & False & 0.743 & 1.224 & 0.408 \\
                           &                        &                        &                        & True  & 0.739 & 1.233 & 0.402 \\ \midrule
\multirow{8}{*}{$\infty$}       & \multirow{4}{*}{False} & \multirow{2}{*}{False} & False                  & True  & 0.768 & 1.174 & 0.421 \\
                           &                        &                        & True                   & True  & 0.753 & 1.177 & 0.393 \\  \cmidrule(r){4-8}
                           &                        & \multirow{2}{*}{True}  & False                  & True  & 0.766 & 1.160 & 0.405 \\
                           &                        &                        & True                   & True  & 0.741 & 1.163 & 0.392 \\  \cmidrule(r){3-8}
                           & \multirow{4}{*}{True}  & \multirow{2}{*}{False} & False                  & True  & 0.764 & 1.188 & 0.395 \\
                           &                        &                        & True                   & True  & 0.751 & 1.192 & 0.394 \\ \cmidrule(r){4-8}
                           &                        & \multirow{2}{*}{True}  & False                  & True  & 0.778 & 1.193 & 0.394 \\
                           &                        &                        & True                   & True  & 0.723 & 1.155 & 0.397\\ \bottomrule
\end{tabular}
\caption{
\textbf{Ablations of various \newgist{} mask configurations.}
All these masks assume that the gist tokens are spread along the context that is to be compressed. 
\emph{Number of pooling windows} is how far back each gist token can attend: 1 means it can attend to the previous gist token while $\infty$ means that it can attend all the way to the beginning of the sequence. 
\emph{Contexts attend to gists} indicates whether the non-gist context tokens can attend to previous gist tokens. 
\emph{Gists attend to gists} indicates whether gist tokens can attend to previous gist tokens. 
\emph{Gists attend to self} indicates whether gist tokens can self-attend. 
\emph{Gists attend to BOS} indicates whether we ensure that each gist can always attend to the BOS token.
Note that for $\infty$ pooling windows, that is always the case. 
The model used for these ablations is \newgist{} with the mask modified accordingly and for $5\times$ compression rate. 
We have evaluated all mask configurations on \drop{}, \fairytale{} and \squad{}.
The mask used throughout the paper is highlighted in bold.
}
\label{tab:mask_ablations}
\end{table}

\begin{table}
    \centering
    \includegraphics[trim={0.5cm, 11cm, 6cm, 1.4cm},clip,width=0.88\textwidth]{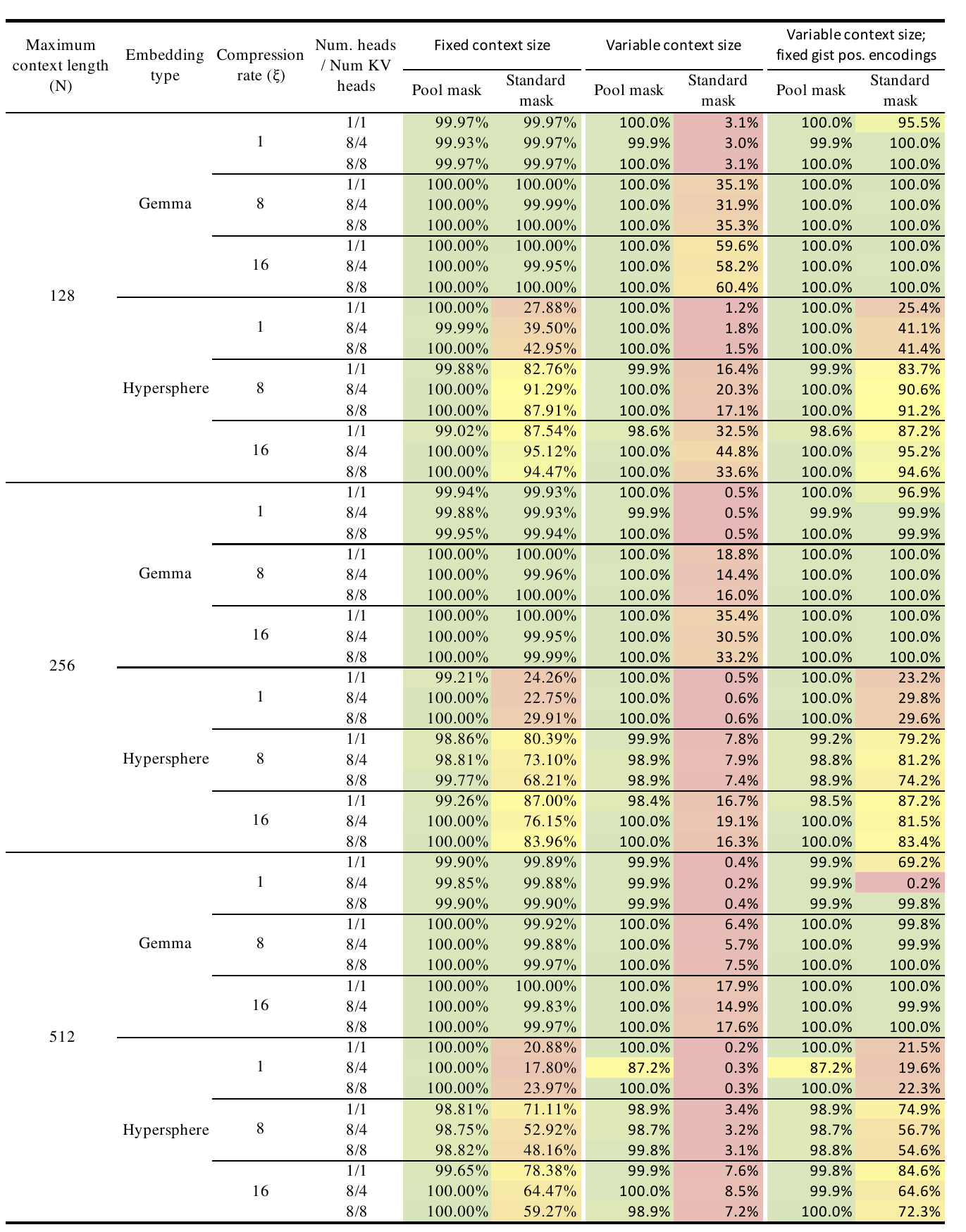}
    \caption{\textbf{Results of the single layer mean pooling experiments in \Cref{sec:single_layer_experiments}.} In order to be able to compare different configurations, we report the fraction of predicted pooled embeddings which correspond to their nearest neighbour in the ground truth mean pooled embeddings. Pool mask ---which ensures that each gist token can only attend to its pooling window--- unsurprisingly achieves near perfect performance across all settings. For the standard causal mask, the results are more varied. When the training and evaluation sample have a fixed length, the transformer layer can learn to mean pool \gemmaemb{} but not \sphereemb{}. When we vary the length of the samples, the model fails to learn any mean pooling with the standard mask. We hypothesized that part of the problem could be that in the variable case, the positional encoding of the gist tokens vary from sample to sample and that interferes with them learning a precise query vector. However, while fixing the positional encodings at the gist locations does improve the performance compared to the variable length case, it still does not match the performance of the pool mask. Therefore, it appears that explicitly restricting the attention as in \newgist{} is needed for focusing the attention.}
    \label{app:single_layer_experiments_results}
\end{table}

\FloatBarrier
\section{Gemini Judge}
\label{app:gemini_judge}

To evaluate the correctness of the model responses, we use an larger model with access to the full context and the correct response to evaluate whether the answer of the model under test is correct.
We used Gemini 1.5 Flash for that purpose with its structured output option.
As sometimes the model would not return a correctly formatted JSON (most often missing the \texttt{original\_response} or \texttt{justification} fields), we retry up to 3 times.
We observed only a few occasions on which the model failed to produce a correctly formatted response after the third attempt.
The prompt used to evaluate all tasks is provided below:

\begin{lstlisting}
You act as a teacher grading a reading comprehension assignment for a student. You will see a long text piece. There is a question following the text. There is also an example of a correct answer provided. You will see the student response. Given the context of the text and the question asked, you are to determine whether the student's response is correct and to provide a justification for your decision. Format the answer in JSON list with dictionaries with two elements: repeat the original student response (a str), correct (a bool) and justification (a str). The justification should be a short sentence explaining why the student's response is correct or incorrect, referring to the student's response, the context of the text, the question and the sample answer.

Here is an example:
Two horses met in the forest. The blue horse had a backpack and said 'Hi'. The red horse wore a hat and said 'Hello'.

Question:
Who had a backpack?

Example correct response:
The blue horse.

Student responses:
The red horse.

The desired output thus is:
(
  "original_response": "The red horse.",
  "correct": false,
  "justification": "The red horse wore a hat, not a backpack."
)

Now it is your turn to grade the student work for this new text.

The text provided to the student is:
{TEXT}

Question:
{QUESTION}

Example correct response:
{SAMPLE_ANSWER}

Student response:
{STUDENT_RESPONSE}
\end{lstlisting}

\section{Gemini Compress}
\label{app:gemini_compress}

\begin{table}
\centering
\begin{tabular}{@{}crrrrrr@{}}
\toprule
Approximate compression rate ($\xi$)& \alpaca{} & \squad{} & \drop{} & \race{} & \narrative{} & \fairytale{} \\ \midrule
2  & 43.4\% & 9.1\%  & 33.7\% & 20.8\% & 16.1\% & 51.5\% \\
5  & 52.5\% & 17.7\% & 36.3\% & 22.7\% & 26.3\% & 57.8\% \\
10 & 64.2\% & 31.5\% & 46.1\% & 28.7\% & 40.6\% & 61.2\% \\
25 &        &        &        &        &        & 67.8\% \\
50 &        &        &        &        &        & 80.0\% \\ \bottomrule
\end{tabular}
\caption{\textbf{Percentage of wrong answers (Gemini Score) for Gemini Compress.} We report the Gemini Score for all 6 datasets when Gemini is prompted to compress the context. As explained in \Cref{app:gemini_compress}, the compression rate is approximate as Gemini cannot exactly adhere to the length constraints.}
\label{tab:gemini_compress}
\end{table}

In order to compare the performance of our models, we use a simple baseline where we ask Gemini model itself to compress the samples using a given compression rate. This is the prompt we used to ask the model to produce such a compression:

\begin{lstlisting}
Compress the following text to be {PERCENT}%
Character count includes all characters, spaces, punctuation.
The original number of characters in the text is {NUM_CHAR}.
The desired number of characters should be around {DESIRED_CHAR}.
Maximize information density above all else.
Retain key information about instruction.
Omit readability, narrative, explanatory wording, and unnecessary punctuation.
Output only the compressed text that has approximately {DESIRED_CHAR} characters (+/- 10 characters).

Text to compress:
{TEXT}
\end{lstlisting}

After the compressed text is produced, we prompt the model to answer questions using only compressed representation using the following prompt:

\begin{lstlisting}
Answer the question *factually* using *only* the provided compressed text, as if it were the complete original source.  Output *only* the direct answer.

Compressed context:
{COMPRESSED_CONTEXT}

Question about the context:
{QUESTION}
\end{lstlisting}

Finally, we use \geminijudge{} as defined in \Cref{app:gemini_judge} to evaluate the answer quality.

\begin{figure}
    \centering
    \includegraphics[width=0.8\columnwidth]{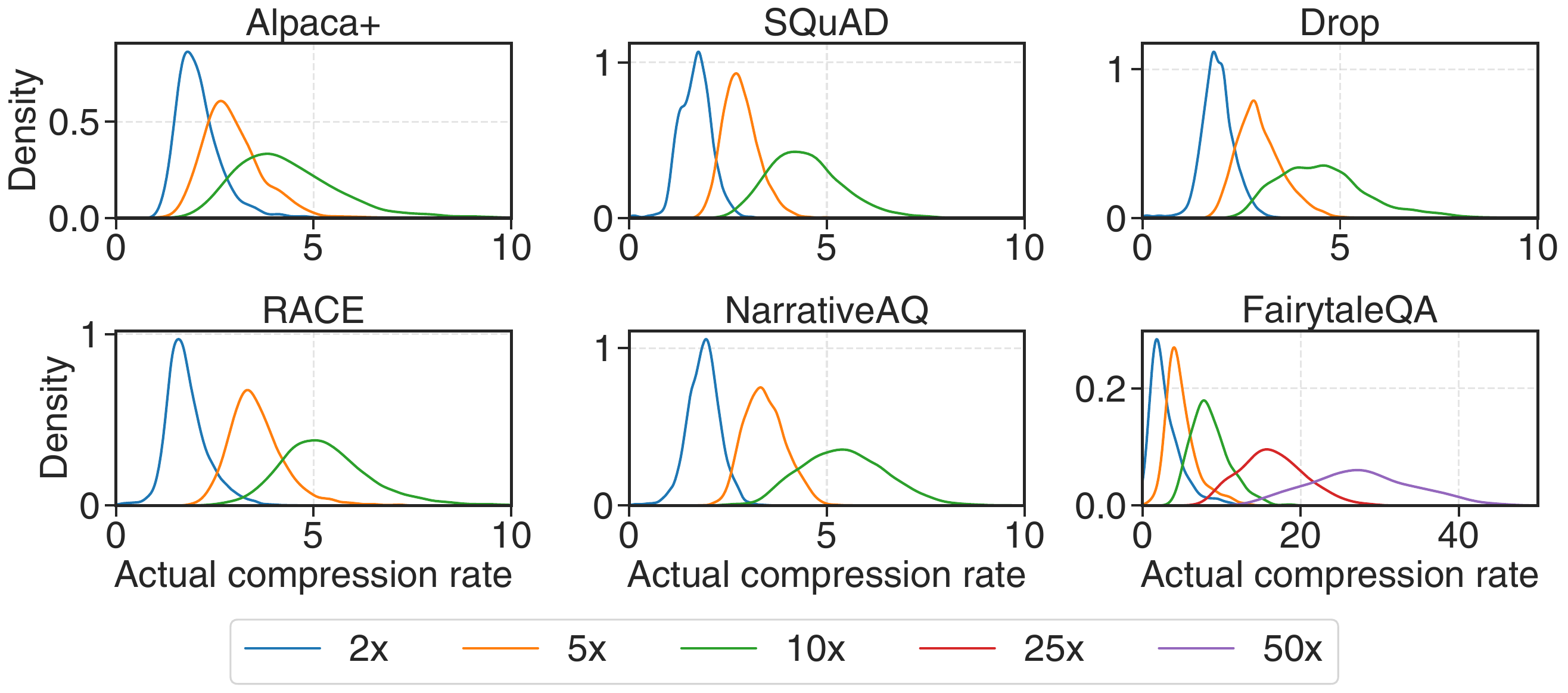}
    \caption{\textbf{Comparison of desired vs.\ actual context compression rates using \geminicompress{}.} We do not have a direct way to control the length of the generated summaries, hence they have a distribution of lengths rather than fixed lengths. Nevertheless, the mean compression rates are close to the target compression rates. }
    \label{fig:gemini_compress_rate}
\end{figure}

In reality, the compressed representation almost never has the desired compression rate, despite our best efforts with prompting. Actual compression rate is about half of what the desired compression rate should be, especially for high compression rates (see \Cref{fig:gemini_compress_rate}).
The results of using Gemini Compress can be seen in \Cref{tab:gemini_compress}.

\subsection{Example of \geminicompress{} on \drop{} dataset}

Here we provide an example of from \drop{} dataset along with compressed representations generated by \geminicompress{} using various compression rates.

\begin{minipage}[c]{\textwidth}
\begin{lstlisting}
CONTEXT:
 According to the market research firm SuperData, as of May 2015, the global games market was worth USD 74.2 billion. By region, North America accounted for $23.6 billion, Asia for $23.1 billion, Europe for $22.1 billion and South America for $4.5 billion. By market segment, mobile games were worth $22.3 billion, retail games 19.7 billion, free-to-play Massively multiplayer online game 8.7 billion, social games $7.9 billion, PC Downloadable content 7.5 billion, and other categories $3 billion or less each.
QUESTION:
Which area accounted for less of the games market, South America or Europe? 
ANSWER: South America
\end{lstlisting}
\end{minipage}

Compression rate $2.2\times$:
\begin{lstlisting}
COMPRESSED_CONTEXT:
SuperData (May 2015): Global games market $74.2B. Region: North America $23.6B, Asia $23.1B, Europe $22.1B, South America $4.5B. Market segment: Mobile $22.3B, Retail $19.7B, F2P MMO $8.7B, Social $7.9B, PC DLC $7.5B, other <=3B.
ANSWER: South America
CORRECT: True
\end{lstlisting}

Compression rate $3.17\times$:
\begin{lstlisting}
COMPRESSED_CONTEXT:
May 2015 global games market $74.2B. Regions: NA $23.6B, Asia $23.1B, Europe $22.1B, SA $4.5B. Segments: Mobile $22.3B, Retail $19.7B, F2P MMO $8.7B.
ANSWER: South America
CORRECT: True
\end{lstlisting}

Compression rate $5.2\times$:
\begin{lstlisting}
COMPRESSED_CONTEXT:
2015: Global game market $74.2B. NA $23.6B, Asia $23.1B, EU $22.1B. Mobile $22.3B, Retail $19.7B.
ANSWER: The provided context does not mention South America. Europe accounted for $22.1B.
CORRECT: False
\end{lstlisting}

\section{Samples from the datasets}
\label{app:dataset_samples}
\subsection{Samples from \alpaca{}}

\begin{minipage}[c]{\textwidth}
\begin{lstlisting}
INSTRUCTION: Give three tips for staying healthy.
INPUT: None
OUTPUT:
1.Eat a balanced diet and make sure to include plenty of fruits and vegetables. 
2. Exercise regularly to keep your body active and strong. 
3. Get enough sleep and maintain a consistent sleep schedule.
\end{lstlisting}
\end{minipage}

\begin{minipage}[c]{\textwidth}
\begin{lstlisting}
INSTRUCTION: Explain why the following fraction is equivalent to 1/4
INPUT: 4/16
OUTPUT: 
The fraction 4/16 is equivalent to 1/4 because both numerators and denominators are divisible by 4. Dividing both the top and bottom numbers by 4 yields the fraction 1/4.
\end{lstlisting}
\end{minipage}

\begin{minipage}[c]{\textwidth}
\begin{lstlisting}
INSTRUCTION: Classify the following into animals, plants, and minerals
INPUT: Oak tree, copper ore, elephant
OUTPUT: 
Oak tree: Plant
Copper ore: Mineral
Elephant: Animal
\end{lstlisting}
\end{minipage}

We format the \alpaca{} samples as such:

\begin{minipage}[c]{\textwidth}
\begin{lstlisting}
Instruction: {INSTRUCTION} Input: {INPUT} Response: {OUTPUT}
\end{lstlisting}
\end{minipage}

Some \alpaca{} samples do not have an \texttt{INPUT} field. We format these samples as:

\begin{minipage}[c]{\textwidth}
\begin{lstlisting}
Instruction: {INSTRUCTION} Response: {OUTPUT}
\end{lstlisting}
\end{minipage}

\subsection{Samples from \squad{}}

\begin{minipage}[c]{\textwidth}
\begin{lstlisting}
CONTEXT:
The difference in the above factors for the case of theta=0 is the reason that most broadcasting (transmissions intended for the public) uses vertical polarization. For receivers near the ground, horizontally polarized transmissions suffer cancellation. For best reception the receiving antennas for these signals are likewise vertically polarized. In some applications where the receiving antenna must work in any position, as in mobile phones, the base station antennas use mixed polarization, such as linear polarization at an angle (with both vertical and horizontal components) or circular polarization.	
QUESTION:
What is one use that would require an antenna to receive signals in various ways at once?	
ANSWER: mobile phones
\end{lstlisting}
\end{minipage}

\begin{minipage}[c]{\textwidth}
\begin{lstlisting}
CONTEXT:
Plant responses to climate and other environmental changes can inform our understanding of how these changes affect ecosystem function and productivity. For example, plant phenology can be a useful proxy for temperature in historical climatology, and the biological impact of climate change and global warming. Palynology, the analysis of fossil pollen deposits in sediments from thousands or millions of years ago allows the reconstruction of past climates. Estimates of atmospheric CO2 concentrations since the Palaeozoic have been obtained from stomatal densities and the leaf shapes and sizes of ancient land plants. Ozone depletion can expose plants to higher levels of ultraviolet radiation-B (UV-B), resulting in lower growth rates. Moreover, information from studies of community ecology, plant systematics, and taxonomy is essential to understanding vegetation change, habitat destruction and species extinction.	
QUESTION:
How can climate changes be determined from soil?
ANSWER: fossil pollen deposits in sediments
\end{lstlisting}
\end{minipage}

\begin{minipage}[c]{\textwidth}
\begin{lstlisting}
CONTEXT:
Situated on one of the world's largest natural harbors, New York City consists of five boroughs, each of which is a separate county of New York State. The five boroughs - Brooklyn, Queens, Manhattan, the Bronx, and Staten Island - were consolidated into a single city in 1898. With a census-estimated 2014 population of 8,491,079 distributed over a land area of just 305 square miles (790 km2), New York is the most densely populated major city in the United States. As many as 800 languages are spoken in New York, making it the most linguistically diverse city in the world. By 2014 census estimates, the New York City metropolitan region remains by a significant margin the most populous in the United States, as defined by both the Metropolitan Statistical Area (20.1 million residents) and the Combined Statistical Area (23.6 million residents). In 2013, the MSA produced a gross metropolitan product (GMP) of nearly US$1.39 trillion, while in 2012, the CSA generated a GMP of over [...]
QUESTION:
What is the size of New York City in square miles?	
ANSWER: 305
\end{lstlisting}
\end{minipage}

We format the \squad{} samples as such:
\begin{lstlisting}
Background information: {CONTEXT} Question: {QUESTION} Answer: {ANSWER}
\end{lstlisting}

\subsection{Samples from \drop{}}

\begin{minipage}[c]{\textwidth}
\begin{lstlisting}
CONTEXT:
As of the census of 2000, there were 120,546 people, 41,668 households, and 32,292 families residing in the county. The population density was 262 people per square mile (101/km2). There were 43,903 housing units at an average density of 95 per square mile (37/km2). The racial makeup of the county was 68.51%
QUESTION:
How many more people are there than families?	
ANSWER: 88254
\end{lstlisting}
\end{minipage}

\begin{minipage}[c]{\textwidth}
\begin{lstlisting}
CONTEXT:
The Mongols' greatest triumph was when Kublai Khan established the Yuan dynasty in China in 1271. The Yuan dynasty created a "Han Army" out of defected Jin troops and an army of defected Song troops called the "Newly Submitted Army" . The Mongol force which invaded southern China was far greater than the force they sent to invade the Middle East in 1256. The Yuan dynasty established the top-level government agency Bureau of Buddhist and Tibetan Affairs to govern Tibet, which was conquered by the Mongols and put under Yuan rule. The Mongols also invaded Sakhalin Island between 1264 and 1308. Likewise, Korea became a semi-autonomous vassal state and compulsory ally of the Yuan dynasty for about 80 years. The Yuan dynasty was eventually overthrown during the Red Turban Rebellion in 1368 by the Han Chinese who gained independence and established the Ming dynasty.	
QUESTION:
Which army did the defected Song troops join, the Han Army or the Newly Submitted Army?	
ANSWER: Newly Submitted Army
\end{lstlisting}
\end{minipage}

\begin{minipage}[c]{\textwidth}
\begin{lstlisting}
CONTEXT:
After defeating the Redskins at home, the Rams traveled on the road against the Bears. The Bears scored first in the first quarter with a 54-yard field goal from Robbie Gould to take a 3-0 lead for the only score of the quarter. In the 2nd quarter, the Bears increased their lead when Michael Bush scored a touchdown on a 5-yard run to make the score 10-0. The Rams responded with Greg Zuerlein's 56-yard field goal to shorten the lead to 10-3 at halftime. In the 3rd quarter, the Rams drew closer as Zuerlein kicked a 46-yard field goal to make the score 10-6 for the only score of the quarter. But in the 4th quarter, the Bears held on for victory as Gould kicked a 22-yard field goal to make it 13-6 and then on the Rams' next possession, Sam Bradford was intercepted by Major Wright who then returned it 45 yards for a touchdown to make it 20-6. Gould kicked a 37-yard field goal to make the final score 23-6 as the Rams dropped to 1-2.	
QUESTION:
How many yards were each of Greg Zuerlein's field goals?	
ANSWER: 56-yard
\end{lstlisting}
\end{minipage}

We format the \drop{} samples as such:
\begin{lstlisting}
Background information: {CONTEXT} Question: {QUESTION} Answer: {ANSWER}
\end{lstlisting}

\subsection{Samples from \narrative{}}

\begin{lstlisting}
CONTEXT:
Several weeks after returning to Kansas from the Land of Oz, Dorothy Gale looks out of her bedroom window and sees a bright and beautiful rainbow on the horizon. She notices that the rainbow is approaching her and Toto as both of them run towards it. Dorothy starts to see Glinda the Good Witch who tells Dorothy that she must return to Oz so that she can save Scarecrow, Tin Man, and Cowardly Lion. Dorothy and Toto reclaim the silver shoes as they find a note from Glinda and Princess Ozma stating that the silver shoes can take her to the Land of Oz and back for the Impassable Desert has taken away much of their power.Dorothy and Toto arrive in the Land of Oz where the items that Dorothy has in her pocket are a small mirror, a safety pin, a glass bottle, and four of Aunt Em's home-made oatmeal cookies. Dorothy and Toto were wondering which direction should they take when they encounter a molasses-covered owl named Wiser. Wiser tells Dorothy that she is in Gillikin Country and tells her to head to Candy County and ask the Great Royal Marshmallow that rules over Candy Country. Arriving at Princess Gayelette's palace, Dorothy and Toto encounter the castle's Jester who welcomes them. The Jester tells them that Princess Gayelette and Prince Quelala have gone missing, adding that they disappeared during a party at the palace which had become haunted and points them in the direction of the castle. When Dorothy and Toto enter the palace, they find a wand that belonged to the Wicked Witch of the West lying on the table.Dorothy reminds the Jester that jesters are supposed to make people happy causing the Jester to freeze in his tracks as the Wicked Witch of the West's ghost urges the Jester to turn Dorothy into a china doll. The Jester gives up the wand as the Wicked Witch of the West's ghost fades away. Thus, the spell is broken and everyone is returned to normal. Scarecrow, Tin Man, Cowardly Lion, and Toto rejoice now that the spell is broken. When the Cowardly Lion asks Dorothy on what she plans to do with the Wicked Witch of the West's wand, Scarecrow and Tin Man plan to keep the wand locked up in a case until they can give it to Glinda and Princess Ozma.Dorothy returns to Kansas where they reunited with Aunt Em and Uncle Henry. The three of them then see a rainbow in the twilight sky which Dorothy hasn't seen before. Dorothy knows that is must be Princess Ozma, Glinda, and the Wizard of Oz's way of saying goodbye to her. The rainbow shimmered over the prairie with all the bright and true colors of the Land of Oz.
QUESTION:
Whose wand does Dorothy and Toto see on the table?
ANSWER: Wicked witch of the west
\end{lstlisting}

\begin{minipage}[c]{\textwidth}
\begin{lstlisting}
CONTEXT:
The novel begins with Silas Lapham being interviewed for a newspaper profile, during which he explains his financial success in the mineral paint business. The Lapham family is somewhat self-conscious in their sudden rise on the social ladder and often fumble in their attempts at following etiquette norms. They decide to build a new home in the fashionable Back Bay neighborhood, and Lapham spares no expense ensuring it is at the height of fashion.Tom Corey, a young man from a well-respected high-class family, shows an interest in the Lapham girls; Mr. and Mrs. Lapham assume he is attracted to Irene, the beautiful younger daughter. Corey joins the Lapham's paint business in an attempt to find his place in the world, rather than rely on the savings of his father, Bromfield Corey. When Tom Corey begins calling on the Laphams regularly, everyone assumes his interest in Irene has grown, and Irene takes a fancy to him. Corey, however, astounds both families by revealing his love for Penelope, the elder, more plain-looking, but more intelligent daughter who possesses an unusual sense of humor, a sophisticated literary passion, and a sensible but inquiring mind. Though Penelope has feelings for Tom Corey, she is held back by the romantic conventions of the era, not wanting to act on her love for fear of betraying her sister.Silas Lapham's former business partner Milton K. Rogers reappears in his life, asking for money for a series of schemes. Mrs. Lapham urges her husband to support the man, whom he had pushed out of the paint company in what was deemed an inappropriate manner. Lapham's dealings with Rogers, however, result in a substantial financial loss. His major asset, the new home on Beacon Street, burns down before its completion. The Laphams are humbly forced to move to their ancestral home in the countryside, where the mineral paint was first developed.
QUESTION:
Which of the Lapham girls is Tom really interested in? 
ANSWER: The older Penelope Lapham
\end{lstlisting}
\end{minipage}

We format the \narrative{} samples as such:

\begin{lstlisting}
Background information: {CONTEXT} Question: {QUESTION} Answer: {ANSWER}
\end{lstlisting}

\subsection{Samples from \race{}}

\begin{lstlisting}
CONTEXT:
Americans have always been hungry for the holidays. After all, a big Thanksgiving feast is one of our country's oldest traditions, older than America itself. Thankfully, the spirit behind Thanksgiving has never changed, either. It has always been a special time to be thankful for the blessings of the past year. The feast that has become known as the First Thanksgiving was actually a harvest festival celebrated in December of 1621. That's when English settlers in Plymouth, Massachusetts, gave thanks for the progress they had made after a harsh winter in their new country. Guests at outdoor tables gobbled up ducks, geese turkeys, clams, eels, fish, wild plums, corn bread and other goodies. About 90 Native Americans also came and brought five deer to add to the feast. The festival lasted for three days. Thanksgiving customs spread and expanded along with the rest of America. After the American Revolution, George Washington proclaimed that the first national Thanksgiving would be on November 26, 1789. In the decades to follow, however, people celebrate Thanksgiving locally and with no official date. A women's magazine editor named Sarah Josepha Hale wanted to change this. After years of trying hard to get support, she finally persuaded President Abraham Lincoln to proclaim the last Thursday in November 1863 as a national day of Thanksgiving. It stayed that way for 75 years afterward until 1939, when President Franklin D. Roosevelt set it one week earlier. He wanted to lengthen the shopping period before Christmas to encourage gift-buyers and help businesses. So Congress ruled that, after 1941, Thanksgiving would be an official federal holiday falling each year on the fourth Thursday of November. This year we celebrated Thanksgiving on Thursday, November 26.
QUESTION:
What can we NOT learn from the passage?
(A) Some Native Americans also joined the First Thanksgiving. (B) Americans will have three days off on Thanksgiving Day. (C) Roosevelt set Thanksgiving one week earlier to develop economy. (D) People still celebrated Thanksgiving with no official date in 1809. 
ANSWER: Americans will have three days off on Thanksgiving Day.
\end{lstlisting}

\begin{minipage}[c]{\textwidth}
\begin{lstlisting}
CONTEXT:
My 16-year-old son, Anton, had gone to the local swimming hole. Most of the kids who swim there are fit and strong teens, and there are plenty of rocks for them to use as safe harbors, so I had no fears for his well-being. Still, the firefighter\'s first words, "You need to come up here to the Stillwater River," made me catch my breath. When I got to the river, I saw Anton sitting quietly on a low platform of the fire engine, with a towel wrapped about his shoulders. I hurried over to him. "You OK?" I asked. "Yeah," was all he said. But my eyes begged for an explanation. I didn\'t get it from my son. The story was this: A couple in their 20s, unfamiliar with the Stillwater, had gotten caught in the current and began screaming for help. Without hesitation Anton and his friend dived into the water, swam out to the drowning woman, and brought her safely to shore. In an age in which the world "hero" is broadcast with abandon and seemingly applied to anyone, I realized the real thing in my son and his friend--the disregarding of personal safety for the sake of another human being. I know that teens are headstrong and self-centered, but this didn't lower the gravity of the event and the desire to do good. Along the way home I tried to get some more information from him, but the only words were, "What\'s for supper?" I thought twice about the tragedy that might have been. Questions flew across my mind like a flight of swallows: Would I have risked my life to save a drowning person? Or would I have chosen to dial 911? Would I have told the story over and over to anyone who\'d listen? The next morning, when Anton got up, I half expected him to tell me the story from his point of view, now that he had some distance from the event. But all he did was to toast a pie, pull himself together, and head for the door to begin the new day.
QUESTION:
Anton kept silent about his deed because _ .
(A) he was still in fear (B) he was annoyed with mother (C) he regarded it as a normal thing (D) he was afraid of being scolded
ANSWER: he regarded it as a normal thing
\end{lstlisting}
\end{minipage}

We format the \race{} samples as such:

\begin{lstlisting}
Background information: {CONTEXT} Question: {QUESTION} Answer: {ANSWER}
\end{lstlisting}

\subsection{Samples from \fairytale{}}

\begin{lstlisting}
STORY:
The story takes place in Baghdad during the Abbasid era. Ali Baba and his elder brother Cassim are the sons of a merchant. After the death of their father, the greedy Cassim marries a wealthy woman and becomes well-to-do, building on their father's business - but Ali Baba marries a poor woman and settles into the trade of a woodcutter.
One day Ali Baba is at work collecting and cutting firewood in the forest, and he happens to overhear a group of forty thieves visiting their treasure store. The treasure is in a cave, the mouth of which is sealed by magic. It opens on the words "Open, Simsim", and seals itself on the words "Close, Simsim". When the thieves are gone, Ali Baba enters the cave himself, and takes some of the treasure home.
Ali Baba borrows his sister-in-law's scales to weigh this new wealth of gold coins. Unbeknownst to Ali, she puts a blob of wax in the scales to find out what Ali is using them for, as she is curious to know what kind of grain her impoverished brother-in-law needs to measure. To her shock, she finds a gold coin sticking to the scales and tells her husband, Ali Baba's rich and greedy brother, Cassim. Under pressure from his brother, Ali Baba is forced to reveal the secret of the cave. Cassim goes to the cave and enters with the magic words, but in his greed and excitement over the treasures forgets the magic words to get back out again. The thieves find him there, and kill him. When his brother does not come back, Ali Baba goes to the cave to look for him, and finds the body, quartered and with each piece displayed just inside the entrance of the cave to discourage any similar attempts in the future.
Ali Baba brings the body home, where he entrusts Morgiana, a clever slave-girl in Cassim's household, with the task of making others believe that Cassim has died a natural death. First, Morgiana purchases medicines from an apothecary, telling him that Cassim is gravely ill. Then, she finds an old tailor known as Baba Mustafa whom she pays, blindfolds, and leads to Cassim's house. There, overnight, the tailor stitches the pieces of Cassims' body back together, so that no one will be suspicious. Ali and his family are able to give Cassim a proper burial without anyone asking awkward questions.
The thieves, finding the body gone, realize that yet another person must know their secret, and set out to track him down. One of the thieves goes down to the town and comes across Baba Mustafa, who mentions that he has just sewn a dead man's body back together. Realizing that the dead man must have been the thieves' victim, the thief asks Baba Mustafa to lead the way to the house where the deed was performed. The tailor is blindfolded again, and in this state he is able to retrace his steps and find the house. The thief marks the door with a symbol. The plan is for the other thieves to come back that night and kill everyone in the house. However, the thief has been seen by Morgiana and she, loyal to her master, foils his plan by marking all the houses in the neighborhood with a similar marking.
When the 40 thieves return that night, they cannot identify the correct house and the head thief kills the lesser thief. The next day, another thief revisits Baba Mustafa and tries again, only this time, a chunk is chipped out of the stone step at Ali Baba's front door. Again Morgiana foils the plan by making similar chips in all the other doorsteps. The second thief is killed for his stupidity as well. At last, the head thief goes and looks for himself. This time, he memorizes every detail he can of the exterior of Ali Baba's house.
The chief of the thieves pretends to be an oil merchant in need of Ali Baba's hospitality, bringing with him Forty thieves hiding in oil jarsmules loaded with thirty-eight oil jars, one filled with oil, the other thirty-seven hiding the other remaining thieves. Once Ali Baba is asleep, the thieves plan to kill him. Again, Morgiana discovers and foils the plan, killing the thirty-seven thieves in their oil jars by pouring boiling oil on them. When their leader comes to rouse his men, he discovers that they are dead, and escapes.
To exact revenge, after some time the thief establishes himself as a merchant, befriends Ali Baba's son (who is now in charge of the late Cassim's business), and is invited to dinner at Ali Baba's house. The thief is recognized by Morgiana, who performs a dance with a dagger for the diners and plunges it into the heart of the thief when he is off his guard. Ali Baba is at first angry with Morgiana, but when he finds out the thief tried to kill him, he gives Morgiana her freedom and marries her to his son. Ali Baba is then left as the only one knowing the secret of the treasure in the cave and how to access it. Thus, the story ends happily for everyone except the forty thieves and Cassim.

QUESTION:
What task does Ali Baba entrust to Morgiana?
ANSWER: making others believe that Cassim has died a natural death	
\end{lstlisting}

We format the \fairytale{} samples as such:

\begin{lstlisting}
Story: {STORY} Question: {QUESTION} Answer: {ANSWER}
\end{lstlisting}

\end{document}